\documentclass{article}

\usepackage[preprint]{neurips_2022}

\usepackage[utf8]{inputenc} 
\usepackage[T1]{fontenc}    
\usepackage{hyperref}       
\hypersetup{
    colorlinks = true,
    citecolor = blue,
    linkcolor = magenta}
\usepackage{url}            
\usepackage{booktabs}       
\usepackage{amsfonts}       
\usepackage{nicefrac}       
\usepackage{microtype}      
\usepackage{xcolor}         
\usepackage{bbm}
\usepackage{amsfonts}
\usepackage{booktabs}
\usepackage{url}
\usepackage{nicefrac}
\usepackage{xspace}
\usepackage{multirow}
\usepackage{amsmath}
\usepackage{graphicx}
\usepackage{hyperref}
\usepackage[utf8]{inputenc}
\usepackage{natbib}
\usepackage{xcolor}
\usepackage{ifthen}
\usepackage[normalem]{ulem}
\usepackage{svg}
\usepackage{subfiles}
\usepackage{xspace}
\usepackage{cleveref}
\usepackage{wrapfig}
\usepackage{dialogue}
\usepackage[most]{tcolorbox}
\usepackage{subcaption}
\usepackage{booktabs}
\usepackage{multirow}

\def\Snospace~{\S{}}

\newcommand\eg{e.g.\xspace}
\newcommand\ie{i.e.\xspace}

\newcommand\refsec[1]{\hyperlink{#1}{§\ref{sec:#1}:~\textsc{#1}}}

\newcommand\reffig[1]{Figure~\ref{fig:#1}}

\newcommand\reftab[1]{Table~\ref{tab:#1}}

\ifthenelse{\isundefined{\definition}}{\newtheorem{definition}{Definition}}{}
\ifthenelse{\isundefined{\assumption}}{\newtheorem{assumption}{Assumption}}{}
\ifthenelse{\isundefined{\hypothesis}}{\newtheorem{hypothesis}{Hypothesis}}{}
\ifthenelse{\isundefined{\proposition}}{\newtheorem{proposition}{Proposition}}{}
\ifthenelse{\isundefined{\theorem}}{\newtheorem{theorem}{Theorem}}{}
\ifthenelse{\isundefined{\lemma}}{\newtheorem{lemma}{Lemma}}{}
\ifthenelse{\isundefined{\corollary}}{\newtheorem{corollary}{Corollary}}{}
\ifthenelse{\isundefined{\alg}}{\newtheorem{alg}{Algorithm}}{}
\ifthenelse{\isundefined{\example}}{\newtheorem{example}{Example}}{}

\newcommand\EG{\textbf{Ecosystem Graphs}\xspace}
\newcommand\releasedate{March 16, 2023\xspace}
\newcommand\websiteURL{\url{https://crfm.stanford.edu/ecosystem-graphs}\xspace}
\newcommand\githubURL{\url{https://github.com/stanford-crfm/ecosystem-graphs}\xspace}
\newcommand\assetlongURL[1]{\url{https://crfm.stanford.edu/ecosystem-graphs/index.html?asset=#1}}
\newcommand\assetURL[1]{\href{https://crfm.stanford.edu/ecosystem-graphs/index.html?asset=#1}{#1}}

\newcommand\numnodes{{262}\xspace}
\newcommand\numedges{{356}\xspace}
\newcommand\numdocumentationentries{{3850}\xspace}
\newcommand\numdatasets{{64}\xspace}
\newcommand\nummodels{{128}\xspace}
\newcommand\numapplications{{70}\xspace}
\newcommand\numorganizations{{63}\xspace}
\newcommand\nummodalities{{9}\xspace}

\title{Ecosystem Graphs: \\
The Social Footprint of Foundation Models}

\author{%
Rishi Bommasani\thanks{Corresponding author.} \\
Stanford University \\
\texttt{nlprishi@stanford.edu} \\
\And
Dilara Soylu \\
Stanford University \\
\texttt{soylu@stanford.edu} \\
\And
Thomas I. Liao \\
Anthropic \\
\texttt{thomas@anthropic.com} \\
\And
Kathleen A. Creel \\
Northeastern University \\
\texttt{k.creel@northeastern.edu} \\
\And
Percy Liang \\
Stanford University \\
\texttt{pliang@cs.stanford.edu} \\
}

\begin{document}
\maketitle

\begin{abstract}
Foundation models (\eg ChatGPT, StableDiffusion) pervasively influence society, warranting immediate social attention.
While the models themselves garner much attention, to accurately characterize their impact, we must consider the broader sociotechnical ecosystem.
We propose \EG as a documentation framework to transparently centralize knowledge of this ecosystem.
\EG is composed of \textit{assets} (datasets, models, applications) linked together by \textit{dependencies} that indicate technical (\eg how Bing relies on GPT-4) and social (\eg how Microsoft relies on OpenAI) relationships.
To supplement the graph structure, each asset is further enriched with fine-grained metadata (\eg the license or training emissions).
We document the ecosystem extensively at \websiteURL: as of \releasedate, we annotate \numnodes assets (\numdatasets datasets, \nummodels models, \numapplications applications) from \numorganizations organizations linked by \numedges dependencies.
We show \EG functions as a powerful abstraction and interface for achieving the minimum transparency required to address myriad use cases.
Therefore, we envision \EG will be a community-maintained resource that provides value to stakeholders spanning AI researchers, industry professionals, social scientists, auditors and policymakers.
\end{abstract}
\section{Introduction}
\label{sec:introduction}

Foundation models (FMs) are the defining paradigm of modern AI \citep{bommasani2021opportunities}.
Beginning with language models 
\citep{
devlin2019bert,
brown2020gpt3,
chowdhery2022palm}, 
the paradigm has expanded to
images \citep{chen2020imagegpt, ramesh2021dalle, radford2021clip},
videos \citep{singer2022makeavideo, wang2022internvideo},
code \citep{chen2021codex},
proteins \citep{jumper2021alphafold, verkuil2022esm2},
and more.
Beyond rapid technology development, foundation models have entered broad social discourse \citep{nyt2020, nature2021, economist2022, cnn2023}.
Given their remarkable capabilities, we are witnessing unprecedented adoption:
ChatGPT amassed 100 million users in just 50 days \citep[the fastest-growing consumer application in history;][]{hu2023chatgpt}
and
Stable Diffusion accrued 30k+ GitHub stars in 90 days \citep[much faster than Bitcoin and Spark;][]{appenzeller2022stablediffusion},
As a bottom line, over 200 foundation model startups have emerged, collectively raising \$3.5B as of October 2022 \citep{kaufmann2022scaleindex}.
In fact, the influx of funding continues to accelerate: Character received \$200M from Andreesen Horowitz, Adept received \$350M from General Catalyst, and OpenAI received \$10B from Microsoft just in Q1 of 2023. 

Foundation models are changing society but what is the nature of this impact?
Who reaps the benefits, who shoulders the harms, and how can we characterize these societal changes?
Further, how do trends in research correspond to outcomes in practice (\eg how do emergent abilities \citep{wei2022emergent} influence deployment decisions, how do documented risks \citep{abid2021persistent} manifest as concrete harms)?
Overall, there is pervasive confusion on the status quo, which breeds further uncertainty on how the space of foundation models will evolve and what change is necessary.
Currently, the AI community and broader public tolerate the uncomfortable reality that models are deployed ubiquitously through products yet we know increasingly little about the models, how they were built, and the mechanisms (if any) in place to mitigate and address harm.

    \begin{figure*}
        \centering
        \begin{subfigure}[b]{0.475\textwidth}
            \centering
            \includegraphics[width=\textwidth]{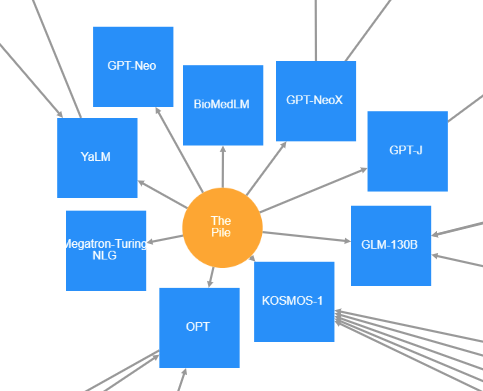}
            \caption{\assetURL{The Pile} dataset \citep{gao2021thepile}}    
        \end{subfigure}
        \hfill
        \begin{subfigure}[b]{0.475\textwidth}  
            \centering 
            \includegraphics[width=\textwidth]{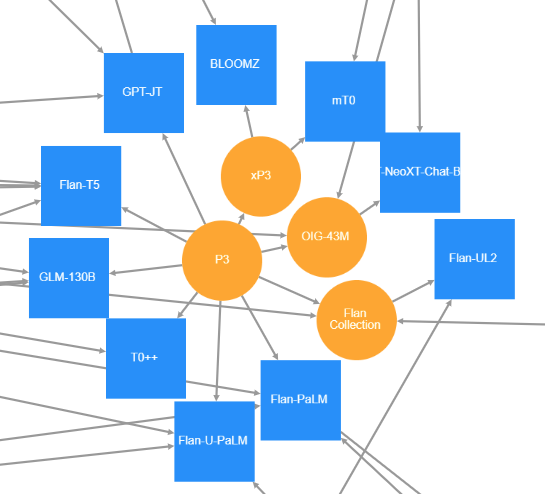}
            \caption{\assetURL{P3} dataset \citep{sanh2021multitask}}
        \end{subfigure}
        \vskip\baselineskip
        \begin{subfigure}[b]{0.475\textwidth}   
            \centering 
            \includegraphics[width=\textwidth]{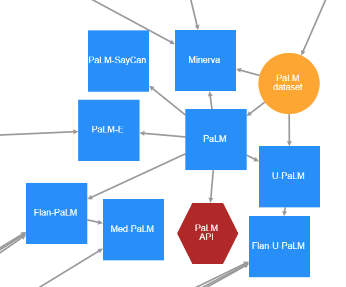}
            \caption{\assetURL{PaLM} model \citep{chowdhery2022palm}}
        \end{subfigure}
        \hfill
        \begin{subfigure}[b]{0.475\textwidth}   
            \centering 
            \includegraphics[width=\textwidth]{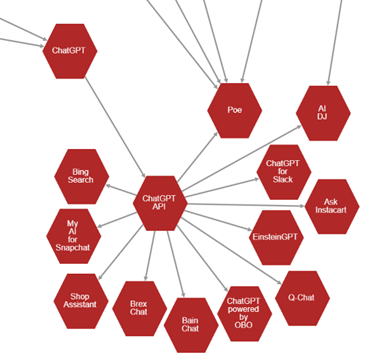}
            \caption{\assetURL{ChatGPT API} \citep{openai2023chatgptapi}}
        \end{subfigure}
\caption{\textbf{Hubs in the ecosystem.} 
To demonstrate the value of \EG, we highlight \textit{hubs}: assets that feature centrally in that many assets directly depend on them.
(a) The Pile is an essential resource for training foundation models from a range of institutions (\eg EleutherAI, Meta, Microsoft, Stanford, Tsinghua, Yandex).
(b) P3 is of growing importance as interest in instruction-tuning grows, both directly used to train models and as component in other instruction-tuning datasets.
(c) PaLM features centrally in Google's internal foundation models for vision (PALM-E), robotics (PaLM-SayCan), text (FLAN-U-PaLM), reasoning (Minerva), and medicine (Med-PaLM), making the recent announcement of an API for external use especially relevant.
(d) The ChatGPT API profoundly accelerates deployment with downstream products spanning a range of industry sectors.
}
\label{fig:hubs}
\end{figure*}


To clarify the societal impact of foundation models, we propose \EG as a centralized knowledge graph for documenting the foundation model \textit{ecosystem} (\reffig{ecosystem-diagram}).
\EG consolidates distributed knowledge to improve the ecosystem's transparency.
\EG operationalizes the insight that significant understanding of the societal impact of FMs is already possible if we centralize available information to analyze it collectively. 

Each node in the graph is (roughly) an \textit{asset} (a dataset, model, or application).
Simply being aware of assets is an outstanding challenge: new datasets are being built, new models are being trained, and new products are being shipped constantly, often with uneven public disclosure.
While attention centers on the foundation model, the technical underpinnings and the social consequences of a foundation model depend on the broader ecosystem-wide context.
To link nodes, we specify \textit{dependencies}: in its simplest form, models require training data and applications require models.
Dependencies are technical relationships between assets  (\eg different ways of training or adapting a foundation model) that induce social relationships between organizations (\eg Microsoft depends on OpenAI because Bing
depends on GPT-4).
Especially for products, surfacing these dependencies
is challenging yet critical: products determine much of the direct impact and dependencies indicate the flow of resources, money, and power.

To supplement the graph structure, we further document each node with an \textit{ecosystem card}, drawing inspiration from other documentation frameworks (\eg data sheets \citep{gebru2018datasheets}, data statements \citep{bender2018data}, model cards \citep{mitchell2018modelcards}).
The ecosystem card contextualizes the node not only in isolation (\eg when was it built), but also with respect to the graph structure (\eg the license affects downstream use, data filters interact with upstream dependencies). 
Documenting applications concretizes societal impact: structural analyses (\eg which organizations wield outsized power) requires grounding out into how people are affected, which is directly mediated by applications. 
We also make explicit new challenges faced in documentation such as 
(i) \textit{maintenance} practices to synchronize the ecosystem graph with the ecosystem,
and (ii) \textit{incentives} that may inhibit or facilitate documentation. 

Given our framework, we concretely document the existing foundation model ecosystem through \numnodes nodes linked by \numedges dependencies and annotated with \numdocumentationentries metadata entries as of \releasedate.
This amounts to \numdatasets datasets (\eg \assetURL{The Pile},  \assetURL{LAION-5B}), \nummodels models (\eg \assetURL{BLOOM}, \assetURL{Make-A-Video}), and \numapplications applications (\eg \assetURL{GitHub CoPilot}, \assetURL{Notion AI}) that span \numorganizations organizations (\eg OpenAI, Google)  and \nummodalities modalities (\eg music, genome sequences).
To briefly demonstrate the value of \EG, we highlight the \textit{hubs} in the graph (\reffig{hubs}), drawing inspiration from the widespread analysis of hubs across other graphs and networks \cite[][\textit{inter alia}]{kleinberg1999hubs, hendricks1995hubs, franks2008extremism, van2013network}.
For asset developers, hubs indicate their assets are high impact; for economists, hubs communicate emergent market structure and potential consolidation of power; for investors, hubs signal opportunities to further support or acquire; for policymakers, hubs identify targets to scrutinize to ensure their security and safety.
In general, \EG functions as a rich interface and suitable abstraction to provide needed transparency on the foundation model ecosystem (\refsec{uses}).
We encourage further exploration at \websiteURL and are actively building \EG by collaborating with the community at \githubURL.
\section{Foundation Model Ecosystem}
\label{sec:ecosystem}

\begin{figure*}
\centering
  \includegraphics[width=\textwidth]{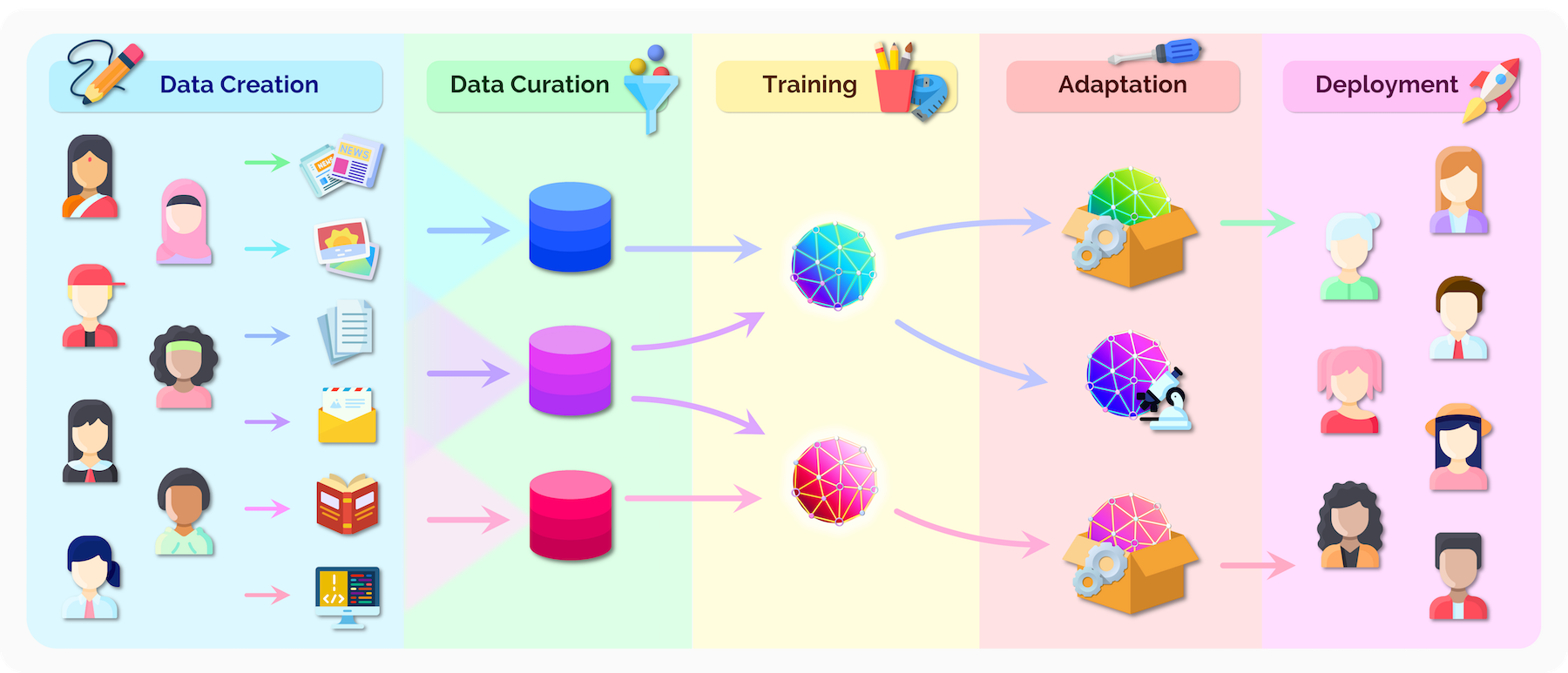}
  \caption{\textbf{Basic foundation model ecosystem.} 
To conceptualize the foundation model ecosystem, we present a simplified pipeline. Image taken from \citet{bommasani2021opportunities}.
  }
  \label{fig:ecosystem-diagram}
\end{figure*}

Technology development situates in a broader social context: technology is the byproduct of organization-internal processes and the artifact of broader social forces \citep{martin2020extending, gebru2018datasheets, mitchell2018modelcards, amironesei2021notes, paullada2021data, birhane2022values}.
\citet[][\S1.2]{bommasani2021opportunities} introduce this perspective for foundation models: \reffig{ecosystem-diagram} demonstrates the canonical pipeline where datasets, models, and applications mediate the relationship between the people on either side.

Concretely, consider \assetURL{Stable Diffusion} \citep{rombach2021highresolution}:
\begin{enumerate}
\item People create content (\eg take photos) that they (or others) upload to the web.
\item LAION curates the LAION-5B dataset \citep{schuhmann2022laion} from the CommonCrawl web scrape, filtered for problematic content. 
\item LMU Munich, IWR Heidelberg University, and RunwayML train Stable Diffusion on a filtered version of LAION-5B using 256 A100 GPUs.
\item Stability AI builds Stable Diffusion Reimagine by replacing the Stable Diffusion text encoder with an image encoder.\footnote{\url{https://stability.ai/blog/stable-diffusion-reimagine}}
\item Stability AI deploys Stable Diffusion Reimagine as an image editing tool to end users on Clipdrop, allowing users to generate multiple variations of a single image (\eg imagine a room with different furniture layouts). 
The application includes a filter to block inappropriate requests and solicits user feedback to improve the system as well as mitigate bias.
\end{enumerate}
This process delineates social roles and, thereby, \textit{stakeholders}: data creators, data curators, compute providers, hardware providers, foundation model developers, downstream application developers, and end users. 
While framed technically (\eg curation, training, adaptaiton), the process bears broader societal ramifications:  for example, ongoing litigation contends the use of LAION and the subsequent image generation from Stable Diffusion is unlawful, infringing on the rights and the styles of data creators.\footnote{See \url{https://www.newyorker.com/culture/infinite-scroll/is-ai-art-stealing-from-artists}.}
As the foundation model ecosystem matures, amounting to greater commercial viability, foundation model development will become even more intricate, implicating even more social roles.\footnote{See \url{https://www.madrona.com/foundation-models/}.}
\section{Documentation Framework}
\label{sec:framework}

To document the foundation model ecosystem, we introduce the \EG framework.
Informally, the framework is defined by a graph comprised of (i) \textit{assets} (\eg ChatGPT), (ii) \textit{dependencies} (\eg datasets used to build to ChatGPT, applications built upon ChatGPT), and (iii) \textit{ecosystem cards} (\eg metadata on ChatGPT). 

\subsection{Definition}
    \begin{figure*}
        \centering
        \begin{subfigure}[b]{0.475\textwidth}
            \centering
            \includegraphics[width=\textwidth]{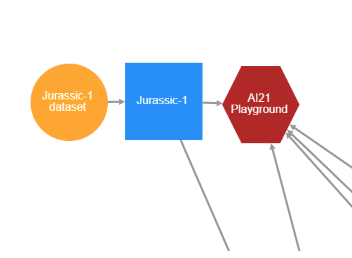}
            \caption{Canonical}    
        \end{subfigure}
        \hfill
        \begin{subfigure}[b]{0.475\textwidth}  
            \centering 
            \includegraphics[width=\textwidth]{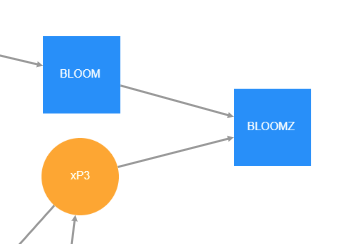}
            \caption{Adaptation}
        \end{subfigure}
        \vskip\baselineskip
        \begin{subfigure}[b]{0.475\textwidth}   
            \centering 
            \includegraphics[width=\textwidth]{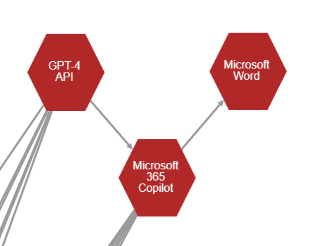}
            \caption{Application layers}
        \end{subfigure}
        \hfill
        \begin{subfigure}[b]{0.475\textwidth}   
            \centering 
            \includegraphics[width=\textwidth]{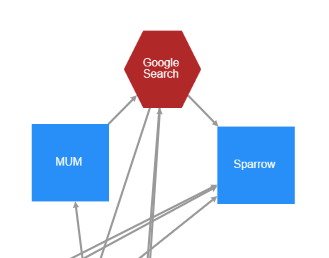}
            \caption{Application dependence}
        \end{subfigure}
\caption{\textbf{Primitive subgraphs in the ecosystem.} 
We spotlight a few 3-node subgraphs to build intuition for \EG.
(a) The standard pipeline: The Jurassic-1 dataset is used to train the Jurassic-1 FM, which is used in the AI21 Playground.
(b) A common adaptation process: BLOOM is adapted by fine-tuning on XP3 to produce the instruction-tuned BLOOMZ model.
(c) Layering of applications: The GPT-4 API powers Microsoft365 Copilot, which is integrated into Microsoft Word.
(d) Dependence on applications: While applications often are the end of pipelines (\ie sinks in the graph), applications like Google Search both depend on FMs like MUM and support FMs (\eg through knowledge retrieval) like Sparrow.
}
\label{fig:primitives}
\end{figure*}

The ecosystem graph is defined in terms of assets: each asset $a \in A$ has a \textit{type} $T(a) \in \{\text{dataset},~\text{model},~\text{application}\}$.
Examples include The Pile dataset, the Stable Diffusion model, and the Bing Search application. 
To define the graph structure, each asset $a$ has a set of dependencies $D(a) \subset A$, which are the assets required to build $a$.
For example, following the Stable Diffusion example in \refsec{ecosystem}, LAION-5B is a dependency for Stable Diffusion and Stable Diffusion is a dependency for Stable Diffusion Reimagine.
Dependencies correspond to directed edges in the ecosystem graph.
In \reffig{primitives}, we given several examples of primitive structures (\ie subgraphs) that we observe in the full ecosystem graph.

To enrich assets with contextual metadata, each asset $a$ is annotated with \textit{properties} $p(a)$ that are stored in the asset's ecosystem card.
Properties include the "organization" that created the asset, the "license" enforced to use the asset, and type-specific properties (\eg the "size" of a model).
For each property (\eg the "license"), we annotate the property as it applies to the asset (\eg the license for Stable Diffusion is the CreativeML OpenRAIL M license).\footnote{\url{https://github.com/CompVis/stable-diffusion/blob/main/LICENSE}} 

\paragraph{Caveats.}
The definition of \EG is deliberately minimalistic: our objective is to make the ecosystem simple to understand to ensure it is legible to diverse stakeholders.
Under the hood, we introduce two additional forms of complexity that we revisit in discussing our implementation of \EG. 
First, while we conceptualize the "nodes" of the ecosystem graph as individual assets, in practice they will instead correspond to sets of closely-related assets (\eg the different model sizes in the GPT-3 model family).
Second, we will annotate properties (\eg the "license") by specifying both a structured \textit{value} (\ie the type of license) and a contextual/qualitative \textit{description} (\eg the provenance for the information). 

Given our definition, we identify five challenges that arise directly from the definition:
\begin{enumerate}
    \item \textbf{Asset discovery.} How do we identify and prioritize the assets?
    \item \textbf{Asset representation.} How do we represent assets?
    \item \textbf{Dependency discovery.} How do we identify the dependencies?
    \item \textbf{Metadata representation.} How do we represent the metadata properties?
    \item \textbf{Metadata annotation.} How do we annotate the metadata for every node?
\end{enumerate}
Beyond this, we should ask who does this work, why they would do it, how the information is maintained (since the ecosystem itself is everchanging), and why it should be trusted.
To ground the \EG framework, we present our concrete implementation before returning to these conceptual challenges and how we navigated them.

\subsection{Implementation}
The ecosystem graph that we have documented thus far is available at \websiteURL.
As of \releasedate, the graph contains \numnodes nodes (\numdatasets datasets, \nummodels models, \numapplications applications) built by \numorganizations organizations that are linked together by \numedges dependencies.

\paragraph{Views.}

\begin{figure*}
\centering
  \includegraphics[width=\textwidth]{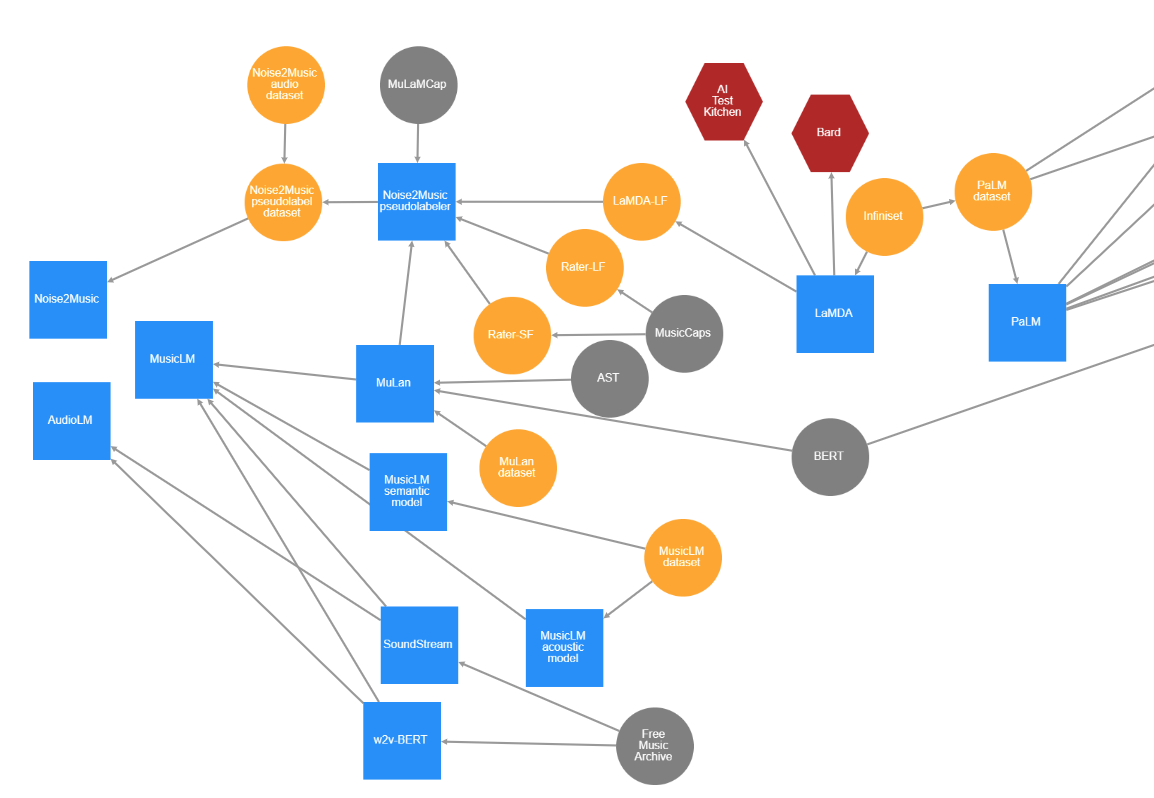}
  \caption{\textbf{Graph view} for \EG as of \releasedate (Google's music-related subgraph).
  We highlight salient foundation models (Noise2Music, AudioLM, MusicLM) as well as the shared and intricate dependencies (\eg on SoundStream and MuLan).
  We also observe that music models at present are often language-controlled (\eg introducing a dependency on LaMDA) that thereby links Google's music models with the more extensive (and more productionized) language models (\eg LaMDA, PaLM).
  More generally, beyond these language-mediated dependencies (\eg outbound dependencies of PaLM), Google's music subgraph is fully contained to the nodes depicted in this figure at present (to our knowledge).
  }
  \label{fig:graph-view}
\end{figure*}

\begin{figure*}
\centering
  \includegraphics[width=\textwidth]{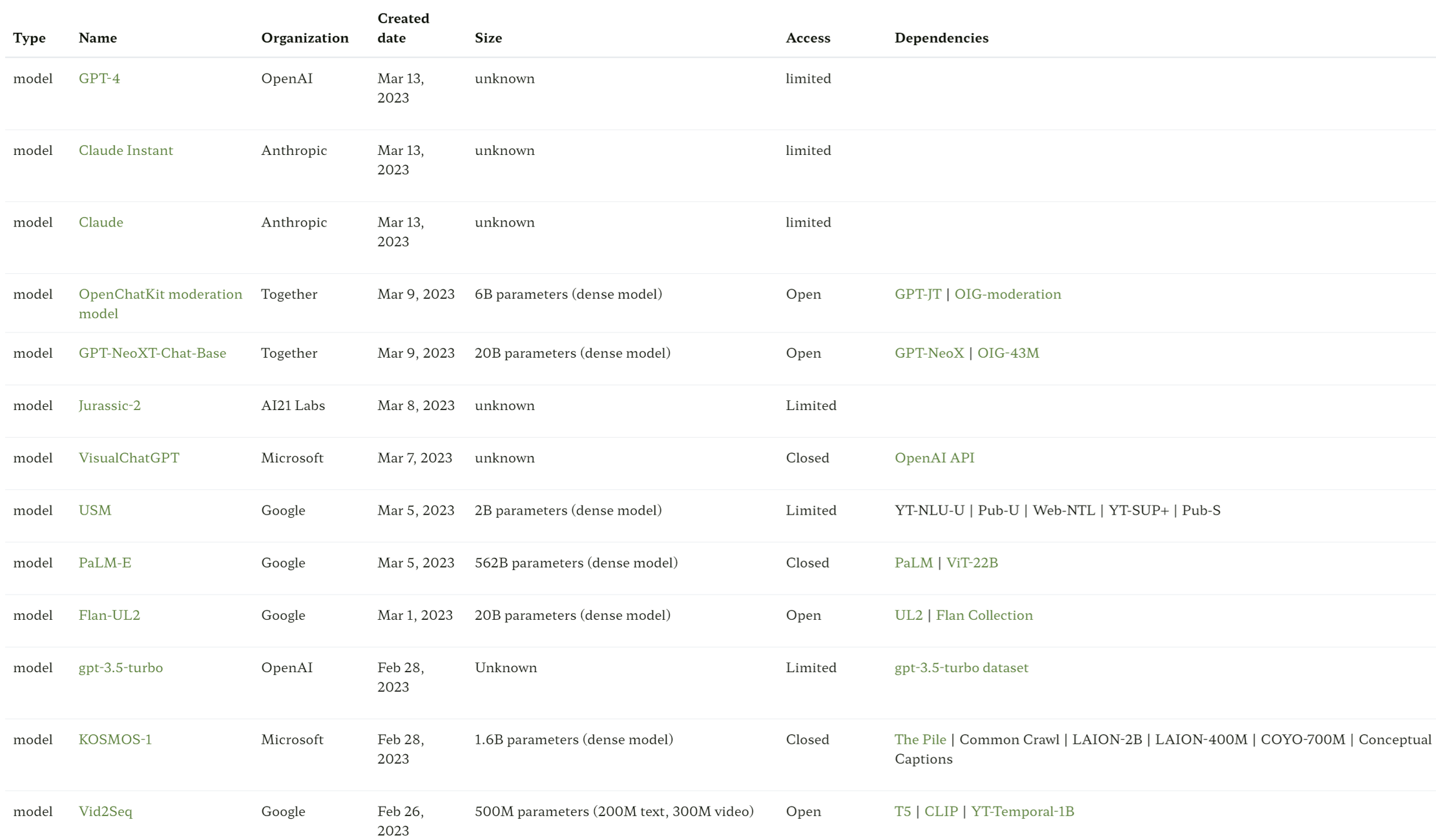}
  \caption{\textbf{Table view} for \EG as of \releasedate (sorted by recent models).
  Over 10 models were released in the 15 day period, with prominent recent models (\eg OpenAI's GPT-4, Anthropic's Claude) disclosing very little to the public (\eg model size, dependency structure). 
  }
  \label{fig:table-view}
\end{figure*}

To visualize the graph structure of the \EG, we provide a simple graph interface shown in \reffig{graph-view}.
Users can zoom into specific regions of the graph to better understand specific subgraphs.
Alternatively, we provide an interactive table (\reffig{table-view}) to search for assets or filter on specific metadata, which can also be exported to a CSV. 
Users can include or exclude specific fields as well as sort by column (\eg the "organization", the asset "type").
Clicking on the node's "name" in either the graph or table views will take the user to the associated ecosystem card.

\paragraph{Ecosystem cards.}

\begin{figure*}
\centering
  \includegraphics[width=0.6\textwidth]{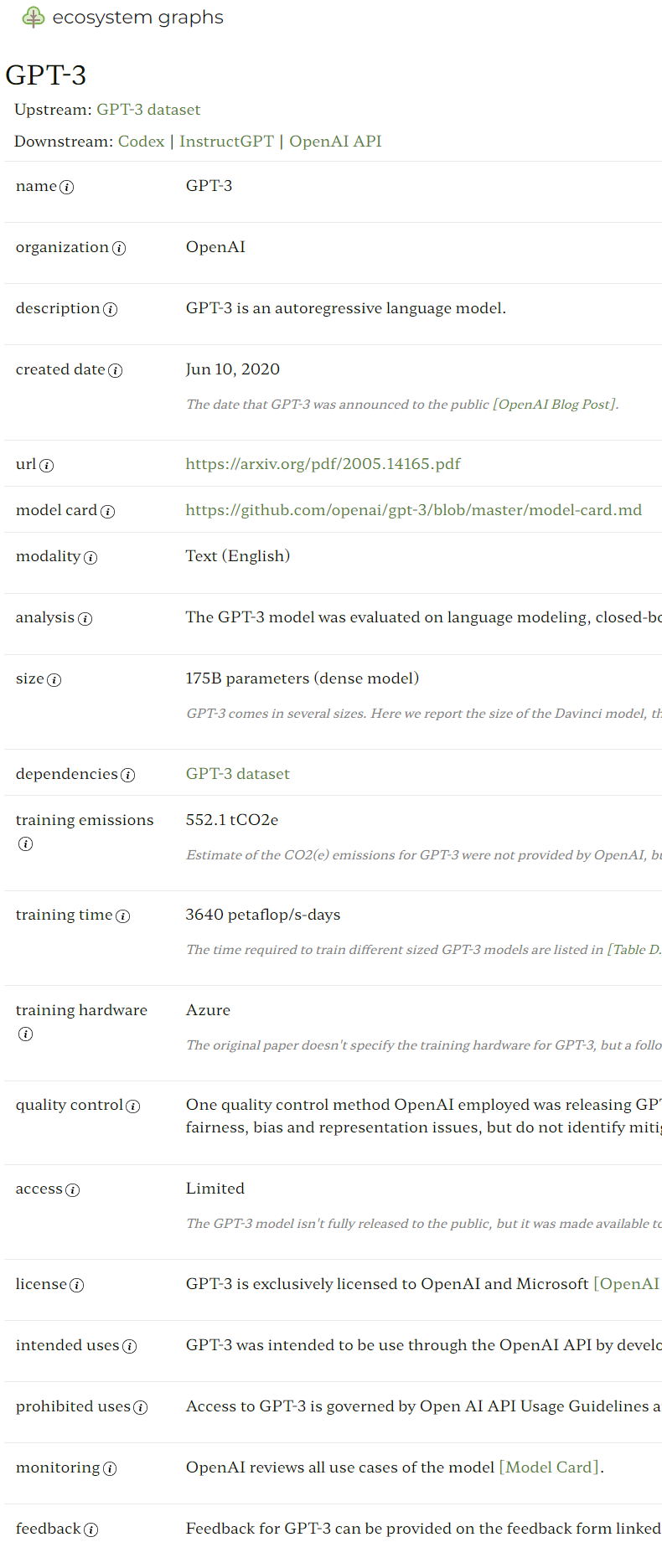}
  \caption{\textbf{GPT-3 Ecosystem Card.} 
  The card contains basic information (\eg that OpenAI developed GPT-3 and when they announced it to the public in 2020), information on how it was built (\eg a standardized statistics on training emissions in tons of C02 emitted and training time in petaflop/s-days), and how it can be built upon (\eg access is available through the OpenAI API and what uses are prohibited by the API usage guidelines). 
  }
  \label{fig:ecosystem-card}
\end{figure*}

Each node is associated with an ecosystem card.
To navigate between adjacent nodes in the graph, the node's dependencies (upstream) and dependents (downstream) are linked to at the top of the page. 
In \reffig{ecosystem-card}, we provide the ecosystem card for GPT-3 from \assetlongURL{GPT-3}: each property includes a help icon that clarifies what the property refers to (see \autoref{tab:all-properties}).
As we describe subsequently, the ecosystem card design aims to centralize useful information, as can be inferred by the abundance of links, rather than replicate the information.

\paragraph{Code.}
On the backend, \EG is a collection of \texttt{YAML} files that store the annotation metadata against a pre-specified schema of fields that matches \reftab{all-properties}.
All aspects of asset selection are handled by the annotator in choosing what to specify in the \texttt{YAML} file: all specified assets are rendered.
For constructing the graph, the dependencies field is used to build edges between the graph: if a dependency is specified but no ecosystem card has been annotated, a stub node is created in the graph.
Anyone can contribute (\eg adding new assets, editing existing assets) by submitting a pull request at \githubURL that will be reviewed by a verified maintainer of the \EG effort. 

\subsection{Assets and nodes}
Assets and nodes are the building blocks for \EG, framing the ecosystem.
We explore how to identify assets (\textit{asset discovery}) and group assets into nodes (\textit{asset representation}).

\paragraph{Asset discovery.}
The value proposition for \EG is to centralize information that was previously distributed: the first step is to be aware of the assets.
For datasets and models that are discussed in research papers, this is relatively straightforward: indeed, many of the foundation models we currently document are the subject of research papers.
However, even for artifacts discussed in papers, we already identify the presence of \textit{dark matter}: assets that must exist, but the public simply knows (essentially) nothing about.
The GPT-4 paper provides a striking example: no details whatsoever are disclosed on the dataset(s) underlying GPT-4 with the report reading "the competitive landscape and the safety implications of large-scale models like GPT-4, this report contains no further details about \dots dataset construction \dots" \citep{openai2023gpt4}.

As we expand our sights to applications and deployment, we encounter further obscurity.
For example, most company-internal foundation models used in products (\eg many of the models used in Google Search) are not disclosed in any regard, let alone the datasets involved in training and improving these models.
And, conversely, we often do not know all the products and applications that depend on publicly-disclosed foundation models (\eg every downstream application built on ChatGPT or Stable Diffusion). 
For this reason, as part of our asset discovery process, we make use of less conventional resources beyond standard academic research: news articles, corporate press releases, venture capital indices, and other heterogeneous sources.
In the future, we may consider automated information extraction from the Internet to keep pace with the scaling of the foundation model ecosystem, though at present the asset discovery process is fully manual and somewhat \textit{ad hoc}.

In discovering assets, we prioritize assets.
While appealing to track every asset, we found this renders the resulting ecosystem graph cumbersome to navigate and introduces untenable annotation burden. 
As a lower bound, there are 150k+ models and 20k+ datasets available on Hugging Face alone.
Consequently, we make our priorities and decisionmaking explicit through principle(s) that influence asset selection.
Namely, we include assets that 
(i) are socially salient, 
(ii) have outsized impact 
or (iii) represent a broader class (\eg ensuring we include at least one music foundation model).
These are imprecise and subjective criteria: we believe all current assets can be justified under them, but we tend to be liberal in our interpretation, preferring to over-include rather than under-include.
As the ecosystem graph expands and its value/uses become more clear, we imagine revisiting these criteria to arrive at more durable and precise principles.

\paragraph{Asset representation.}
Once we have determined the assets to be included in principle, we observe that many assets are very closely connected.
Therefore, pragmatically, we choose to group closely-related assets together into a single node in representing the ecosystem graph.
We believe this representational choice improves the legibility of the ecosystem.
Concretely, we group assets together into a single node when they are presented as belonging to the same category in the view of their developers or other entities in the ecosystem.
As examples, variants of datasets are often referred to using the same term (\eg many datasets are called "English Wikipedia") as are products (\eg the unannounced versions of Google Search). 
Further, models often belong to a shared collective even if they differ (most often in size) such as the four models introduced by \citet{brown2020gpt3} that are collectively referred to as GPT-3. 
There are trade-offs: by collapsing to one node, we potentially obscure slight differences, though if we observe a distinction is relevant, we can choose to dis-aggregate the node back into its constituent assets. 
Overall, similar to how we determine which assets to include, we construct graph nodes by prioritizing usefulness over faithfulness to the true ecosystem.

\subsection{Dependencies}
To define the graph structure, we need to identify the dependencies of each node (\ie specify $D(a)$ for every $a$).
In practice, determining the dependencies of assets introduced in research papers is fairly straightforward, though there are increasingly exceptions (\eg the aforementioned GPT-4 dataset).
More frequently, even for assets described in research, the assets they depend upon themselves may be opaque.
For example, the AI21 Jurassic-1 models are trained on a dataset that is obliquely described in a single sentence: "Our model was trained with the conventional self-supervised auto-regressive training objective on 300B tokens drawn from publicly available resources, attempting, in part, to replicate the structure of the training data as reported in Brown et al. (2020)." \citep{lieber2021jurassic}.
As a result, we have an asset called "Jurassic-1 training dataset" that the Jurassic-1 models depend upon which will be poorly documented (since we only know the dataset size and nothing else). 
But in other cases, especially those involving assets that are commercial products, the dependency structure itself is almost entirely unknown. 

Overall, we found that annotating the dependency structure is complicated in many practical settings, yet essential for structuring the ecosystem.
For example, from the dependencies alone, we come to realize that assets built by 5+ organizations all depend on EleutherAI's The Pile (see \reffig{hubs}).
And it identifies other forms of structure: for example, many more applications today depend on OpenAI language models than Anthropic language models, even though the models benchmark similarly \citep{liang2022helm}, suggesting differential social impact.

\subsection{Ecosystem cards}
\begin{table*}[htp]
\resizebox{\textwidth}{!}{
\begin{tabular}{lll}
\toprule
\textbf{Category} & \textbf{Field} & \textbf{Description} \\
\midrule
\multirow{12}{*}{{Basic}} & name (G) &  Name of the asset (unique identifier). \\ 
& organization (G) &  Organization that created the asset. \\ 
& description (G) &  Description of the asset. \\ 
& created date (G) &  When the asset was created. \\ 
& url (G) &  Link to website or paper that provides a detailed description of the asset. \\ 
& datasheet (D) &  Link to the datasheet describing the dataset. \\ 
& model card (M) & Link to the model card describing the model. \\
& modality (D, M) & Modalities associated with the asset (\eg text, images, videos). \\ 
& output space (A) & Description of the application's output space (\eg generation, ranking, etc.). \\
& size (D, M) & How big the (uncompressed) dataset is. \\ 
& sample (D) & Small sample of content from the dataset. \\ 
& analysis (D, M) & Description of any analysis (\eg evaluation) that was done. \\ 
\midrule
\multirow{8}{*}{{Construction}} & dependencies (G) &  A list of nodes (\eg assets, models, applications) that were used to create this node. \\ 
& quality control (G) &  What measures were taken to ensure quality, safety, and mitigate harms. \\ 
& included (D) & Description of what data was explicitly included and why. \\ 
& excluded (D) & Description of what data is excluded (\eg filtered out) and why. \\ 
& training emissions (M) & Estimate of the carbon emissions used to create the model. \\ 
& training time (M) & How much time it took to train the model. \\
& training hardware (M) & What hardware was used to train the model. \\
& adaptation (A) & How the model was adapted (\eg fine-tuned) to produce the derivative. \\
\midrule
\multirow{10}{*}{{Downstream}} & access (G) &  Who can access (and use) the asset. \\ 
& license (G) &  License of the asset. \\ 
& intended uses (G) &  Description of what the asset can be used for downstream. \\ 
& prohibited uses (G) &  Description of what the asset should not be used for downstream. \\ 
& monitoring (G) &  Description of measures taken to monitor downstream uses of this asset. \\ 
& feedback (G) &  How downstream problems with this asset should be reported. \\ 
& terms of service (A) & Link to the terms of service. \\ 
& monthly active users (A) & Rough order of magnitude of number of active users. \\ 
& user distribution (A) & Demographic and geographic distribution of users. \\ 
& failures (A) & Description of known failures/errors. \\ 
\bottomrule 
\end{tabular}}
\caption{\textbf{Ecosystem card properties.} 
For nodes, we annotate these properties: (G) indicates \textbf{G}eneral properties across all asset types, whereas D/M/A indicate type-specific properties for \textbf{D}atasets/\textbf{M}odels/\textbf{A}pplications, respectively.
}
\label{tab:all-properties}
\end{table*}

Having defined the graph structure, we instrument each node by further documenting properties, drawing inspiration from existing documentation approaches such as data sheets \citep{gebru2018datasheets} and model cards \citep{mitchell2018modelcards}.
We iteratively developed our collection of properties based on two principles:
(i) we emphasize properties that are ecosystem-centric (\eg how nodes are influenced by their dependencies or shape their dependents)
and (ii) we offload documentation that exists elsewhere (\eg pointing to existing model cards) to avoid reinventing the wheel.
For each node, we refer to the associated metadata as the node's \textit{ecosystem card}.

In \reftab{all-properties}, we decompose the processing of filling out the ecosystem card into:
\begin{enumerate}
    \item \textit{Basic} properties of the node (\eg the developer organization).
    \item \textit{Construction} properties of the node (\eg the training emissions for models).
    \item \textit{Downstream} properties of the node (\eg the license and terms-of-service for applications).
\end{enumerate}
For each property, we annotate a \textit{value} and, potentially, a \textit{description} that justifies/explains how the value should be interpreted (\eg attributes a source to provide provenance). 

\subsubsection{Basic properties}
To understand nodes and assets even independent of the broader ecosystem, certain basic information is necessary such as the "name" of the asset(s) and the "organization" that produced the asset(s).
To document these properties proves to be fairly straightforward in practice, though there are some complexities.
Unfortunately, there may be opacity even for these basic properties.  (For example, the naming convention for OpenAI models has been historically unclear\footnote{See \url{https://platform.openai.com/docs/model-index-for-researchers}.}
Or the granularity may not be obvious in the case of organizations: as an example, we annotate Copilot as developed by GitHub even though GitHub has been acquired by Microsoft.
In the case of the "description" field, we generally quote the asset(s) developers, along with providing the "URL" that disclosed the node to the public (\eg the paper or press release). 
In addition to these properties, we specifically highlight the "created date": as \EG is maintained over time, filtering on the date can be used to automatically build timelines and understand how the ecosystem evolves.
As an example, filtering for node(s) before January 1, 2023 vs. after January 1, 2023 makes obvious both (i) the early adopters of foundation models and (ii) how publicly-announced deployments have rapidly accelerated in 2023.

\subsubsection{Construction properties}
A node's dependencies by no means fully determine its nature: there are many choices on how to use these dependencies (\eg many products can be built from the same model, different models can arise from training on the same dataset).
However, it is challenging to meaningfully summarize this: the training procedure for many prominent foundation models can amount to dozens of pages in a well-written paper \citep[see][]{chowdhery2022palm}, meaning even a summary would be very long and likely not much more useful than pointing to the paper itself. 
Having surveyed a variety of assets, we converged to 
(i) a broad umbrella category of \textit{quality control} for all assets, 
(ii) deliberate \textit{inclusion/exclusion} for datasets (\eg filtering out "toxic" content based on a word list, which may have the side-effect of removing LGBTQ+ language \citep{dodge2021c4, gururangan2022whose}), 
(iii) material \textit{training costs} for models (\eg to contextualize environmental impact \citep{lacoste2019quantifying, strubell2019energy, henderson2020towards, luccioni2023counting}) and 
(iv) \textit{adaptation} details for applications (\eg fine-tuning details and UI design). 
We found these details provide important context since dependencies visually look the same in \EG, yet these relationships are non-equal. 

\subsubsection{Downstream properties}
To construct the ecosystem graph, we specify dependencies on the target asset: given a node, we annotate its parents.\footnote{We found this natural since, in general, we may not know the dependents of a given upstream asset (\eg ChatGPT continues to accrue new dependents well after its initial release), but we can better trace the lineage when annotating the downstream asset.} 
However, some properties of an asset influence how it can be built upon, rather than how it was built.
Most notably, the \textit{access} status of an asset directly determines who can build on it, whereas the \textit{intended/prohibited uses} influence how the asset should be built upon (in addition to the \textit{license} and \textit{terms of service}).
In general, we found these properties to be straightforward to annotate, though we find discussion of intended/prohibited uses is quite uneven and in some cases no license/terms of service could be found.

Beyond how the asset is built upon, we further annotate fields that determine the asset's downstream social impact.
Most notably, this makes clear the sense in which applications ground out much of the impact: applications have \textit{end users}.
Further, to build upon the transparency of \EG, we highlight important mechanisms for accountability/recourse that we track: 
(i) can asset developers \textit{monitor} the usage of their assets downstream,
(ii) do specific \textit{failures} or harms concretely arise,
and (iii) when these issues come up, do \textit{feedback} mechanisms exist to propagate this information back upstream?
These fields signal assets that are having high impact, which could confer recognition (or even payment) to those who contributed to their widespread downstream impact (\eg valuing data creators whose data generates value downstream).
 
\subsubsection{Complementarity of construction and downstream properties}
The construction and downstream properties together enrich the underlying graph in an essential way.
When edges are interpreted in the forward direction, they indicate how assets are built; when the edges are interpreted in the backwards direction, they indicate how feedback would flow back upstream.
We stress this point as a reflection of the immaturity of the foundation model ecosystem at present by analogy to other industries.

Concretely, we juxtapose the FM ecosystem with the automobile industry as a more established industry with robust practices for its supply chain.
The National Highway Traffic Safety Administration (NHTSA; an agency under the US Department of Transportation), among other entities, is tasked with ensuring automotive safety.\footnote{See their guidelines on motor vehicle safety at \url{https://www.nhtsa.gov/sites/nhtsa.gov/files/documents/14218-mvsdefectsandrecalls_041619-v2-tag.pdf}.}
Following the forward flow of materials through the supply chain, when a batch of parts (\eg brakes) is found to be sub-standard, established protocols mandate the recall of the fleet of cars built using those parts.
Since the National Traffic and Motor Vehicle Safety Act was enacted in 1966, the NHTSA has recalled over 390 million cars due to safety defects.
Conversely, when several cars (possibly from different manufacturers) are reported to be faulty, the shared source of the defect can be traced by attributing their common sources/parts.
Critically, centralized infrastructure exists for consumers to identify how to report issues to the NHTSA (\eg the Department of Transporation's Vehicle Safety Hotline), to interpret how their report will be used, to understand the investigation process the NHTSA implements, and to understand the legal remedies and consumer protections they are afforded.
And the Federal Motor Vehicle Safety Standards sets formal and transparent standards on what constitutes the minimum performance requirements for each (safety-relevant) automotive part.

In short, the automobile industry and its practices/norms illustrate the virtues of observing the ecosystem as a whole and how both forward and backwards traversals can directly inform action.
If an upstream asset (akin to the faulty brakes) is identified to be faulty in the FM ecosystem, we are not confident that broad communication, let alone interventions like a recall, can reliably be expected to occur.
For example, if LAION-5B was shown to be data poisoned \citep{carlini2023poisoning}, how would the developers of Stable Diffusion, the subsequent application developers who built upon Stable Diffusion, and the broader end userbase be notified?
In part, this uncertainty simply is due to the absence of comparable entities to the NHTSA that are responsible for governing the ecosystem, but also due to the absence of infrastructure for information propagation.
Similarly, many assets themselves lack monitoring mechanisms (especially when assets are released openly, monitoring is currently often nonexistent), let alone publicly-disclosed means for relaying feedback and incident reports upstream.
While we expect the organizations may have partnerships (\eg when Khan Academy users surface issues, Khan Academy and OpenAI likely have significant dialogue to diagnose if these problems arise from Khan Academy's use of OpenAI's GPT-4), we specifically highlight the inadequacy for end users and consumers.
In other words, the basic consumer protections for ensuring an individual adversely affected by a product downstream is able to communicate this upstream and be taken seriously do not exist.

\subsection{Annotation practices}
To annotate each property requires dealing with many types of variation: different assets often have specific properties that are idiosyncratic and, in many instances, information is not publicly available.
As we iterated on annotation best practices, we identified two key concepts: (i) how should we interpret \textit{missing} data entries and (ii) how can we \textit{trust} the recorded information?

\paragraph{Missing data.}
We identified four forms of missing data that arise under different conditions: each form has different semantics, which we clarify by annotating the value for the associated property differently.
\begin{enumerate}
    \item \textbf{None}. A property is annotated with the value \textit{none} if an annotator looked for the information and was unable to find it. For example, for many nodes, we could not find any feedback mechanism or monitoring information. It is possible a feedback form does exist, but given we could not find it, we believe this information is effectively too obscure (\eg unreasonable to expect a consumer to find when reporting feedback).
    \item \textbf{Unknown}. A property is annotated with the value \textit{unknown} if the information must exist (and, potentially, an annotator looked for the information and was unable to find it). For example, the training hardware and training time are often not disclosed for many models, but of course the models were trained on some hardware and took some time to train.
    \item \textbf{Empty string}. A property is annotated with the empty string if the annotator chose to not annotate the property, which generally indicates a lack of time. To ensure all ecosystem cards have non-zero information, we minimally require the basic properties be annotated. 
    \item \textbf{N/A}. A property is annotated with the value \textit{N/A} if the property is not applicable to the node/asset(s). By design, the properties are chosen to be broadly applicable, so we encounter this annotation very rarely.
\end{enumerate}
When aggregated across the entire ecosystem, these conventions for missing data help to articulate pervasive opacity (\ie many \textit{unknown} values) and immaturity (\ie many \textit{none} values) in the foundation model ecosystem. 

\paragraph{Trust.}
To ensure information in \EG is legitimate and credible, we implement two mechanisms.
First, to add or modify information, all such requests must be verified by a vetted maintainer to ensure the correctness of the information as well as the consistency with prior annotations.
Consequently, especially as \EG expands to be a community-maintained artifact, this form of moderation provide some guarantees on the sustained quality of the resource.
In fact, since \EG is implemented as a GitHub repository, the full version history of commits (and any associated discussion) is maintained to ensure the provenance of information (much akin to the Wikipedia change log). 
Moving forward, more sophisticated moderation (\eg akin to Wikipedia maintenance) could be developed.
Second, to source the information, we require that information be attributable to a publicly-available source that is provided in the description accompanying the property's value.
In other words, information should have clear provenance to both ensure the annotation matches the source and that the source itself is reliable.
In the future, we imagine this constraint may prove restrictive (namely because it is inconvenient for reporting privately-disclosed information), but we imagine conventions akin to those in journalism can be adopted if necessary. 

\subsection{Maintenance and Incentives}
Everything in the ecosystem graph is subject to change. 
Since foundation models are being deployed remarkably quickly, keeping pace in documenting the ecosystem is an ongoing challenge.
For example, in the week of March 13, 2023, over a dozen products were announced that all depend on OpenAI's GPT-4.
Further, even for existing assets, their dependencies or metadata may change over time: Anthropic's Claude and Google's PaLM were initially closed access, but now are limited access with the release of their respective APIs for downstream development.

For this reason, we explore \textit{who} will maintain \EG and whether \textit{incentives misalignment} introduces challenges, given much of the value is contingent on the graph being up-to-date and correct.
Moving forward, as foundation models feature more centrally in broader social discourse and \EG sees greater adoption, maintenance could be mandated as a policy requirement to ensure sustained transparency \citep[see][]{bommasani2023transparency, madry2023supplychain}. 

\paragraph{Who maintains the ecosystem graph?}
To this point, the ecosystem graph has been built and maintained by the authors of this work.
Moving forward, while this will continue, this will be increasingly insufficient given the growing scale of the ecosystem.
Therefore, we envision two complementary strategies for expanding the group involved in maintaining \EG. 
Building on traditions of open source software development (\eg Wikipedia, Mozilla, PyTorch, Hugging Face), we actively encourage contributions.
To broaden accessibility, new assets can be submitted from a public Google form to remove the barriers of having a GitHub account and familiarity with GitHub.
Further, for this reason we use a lightweight process with explicit guidelines on how to create and edit entries in \texttt{YAML}.
Drawing upon trends in open source, we will implement processes for top contributors to be recognized for their achievement and signal-boosted in the broader community.\footnote{While we do not currently implement any extrinsic bounties, works like \citet{zhao2017devising} and \citet{chowdhury2021twitter} demonstrate their efficacy, warranting further consideration in the future.}

While open source contributions can be very powerful, permitting decentralized contribution to the shared knowledge resource, they require a culture to form that supports and sustains them.
As \EG grows and is more broadly adopted, we envision it may become a broad-use public repository that organizations themselves are incentivized to maintain.
In the future, we propose that major FM organizations each select a dedicated representative responsible for the upkeep of the organization's nodes (and, possibly, some direct neighbors).
This mechanism introduces \textit{accountability}: the veracity of an organization's nodes and dependencies is the responsibility of this maintainer.
Here, we could lean on practices of periodic public reporting (\eg quarterly financial earnings) in reminding the representative to update the graph on a specific cadence.
We imagine the specifics of this will further sharpen as \EG is more broadly adopted, and as we better understand both the rate of change in the ecosystem and the informational needs that \EG serves. 
In the future, the process of updating the ecosystem graph could be integrated into organizational-internal data entry, since much of what is tracked in \EG is likely already being tracked within organizations. 

\paragraph{The compatibility of incentives.}

Ensuring the ecosystem is transparent serves many informational needs and benefits many stakeholders.
Much akin to other shared knowledge resources (\eg Wikipedia, the US Census), downstream use cases continuously arise, further incentivizing the sustained upkeep of the resource.
In the FM ecosystem specifically, we expect asset developers will be incentivized to disclose the impact of their assets (\eg organizations often put out press releases to disclose information on the widespread use of their products).\footnote{For example, see the strategic partnership between Hugging Face and Amazon: \url{https://huggingface.co/blog/aws-partnership}.}
Or, akin to how food vendors proactively announce their food is organic or how cosmetics companies proactively indicate their cosmetics involve the humane treatment of animals, asset developers should definitely be incentivized to highlight their own responsible conduct.
While incentives may not exist for every bit of information to be made transparent, we hope \EG will encourage increased transparency by demonstrating the value of making information and knowledge public.

However, we do recognize there simultaneously exist pernicious incentives for organizations to maintain opacity in the ecosystem: most directly, transparency could infringe on corporate secrets and commercial interest.
Central to our approach in \EG is recognizing that commercial interest need not entail a blanket ban on transparency: in many cases, information can be made transparent to the public while not compromising any commercial agenda.
In other words, the information in question is common knowledge amidst an organization's competitors, and it is better for the public to have partial transparency rather than to have nothing at all.
This approaches aligns with having representatives of each organization: the process of making assets transparent can involve engaging with the organization, flexibly and iteratively identifying the boundaries of transparency in an organization-specific and asset-specific way (\eg OpenAI's desire for transparency may change from 2021 to 2022, or from CLIP to ChatGPT). 
More expansively, \EG mediates an incremental process for building norms of transparency \citep{liang2022community-norms, bommasani2023transparency} and functions as an inroad for policy intervention as specific informational needs grow more important.
\section{Use Cases for Ecosystem Graphs}
\label{sec:uses}
To demonstrate the value of \EG, we enumerate social roles to explain their concrete informational needs and fundamental questions that \EG addresses.
\EG may be insufficient to fully meet their demand, but we highlight how it makes progress: is a minimum transparency standard and unifying abstraction across diverse use cases.

\paragraph{Foundation model developers.}
FM developers need to be aware of the available assets they can build upon (\eg new instruction-tuning datasets like P3), the assets they will compete with (\eg new releases from competitors), and current/foreseeable demand from downstream application developers.
Each of these informational needs are directly addressed from the ecosystem graph.
In fact, at a more fine-grained level, the ecosystem graph allows developers to compare their practices with their competitors.
For example, developer A may come to realize that developer B implements a more robust quality control pipeline.
While developers likely already track this information themselves about their competitors, centralized information could inform more intentional decisions in choosing to either conform to or diverge from the practices of other developers \citep[\eg norms on release;][]{liang2022community-norms}.

\paragraph{Application developers.}
Akin to foundation model developers, downstream application developers also should be aware of the full space of foundation models in choosing which foundation model to build upon (or investing to build foundation models themselves).
However, the graph structure provides further clarity, since implicitly it indicates which foundation models are more popular (perhaps because they are easier or preferable to build upon). 
This signal is complementary to other signals \citep[\eg comparing models on standardized evaluations like HELM;][]{liang2022helm}, because application developers can make decisions informed by prior developer choices.
For example, several organizations (\ie OpenAI, Cohere, Anthropic) currently offer similar language model API offerings but they have been differentially adopted by downstream developers. 
We could imagine that a new copywriting startup might find this information illuminating in making decisions on which API to use: (i) if they use the same API as their competitors, how will they distinguish themselves, or (ii) if they use a different API, why is it better for them given their competitor chose to use a different one. 
In fact, metadata in the ecosystem card could be composed to further constrain the search space: for example, one could look for a permissively-licensed available model that has been evaluated on relevant benchmarks (by filtering on the "license", "access", and "analysis" properties).

\paragraph{End users of FM applications.}
Consumers deserve to know how the technology they use was built, akin to requirements that require food in the US be labeled with ingredients.\footnote{\url{https://www.fda.gov/food/food-ingredients-packaging/}}
The graph structure and the simple web interface make this practical: a user can look up the product, see if it is in the graph, and trace its dependencies.
This process surfaces any existing mechanisms for feedback reporting, which could prove to be useful if the user experiences an issue.
While the existing documentation is quite scarce, the user would also be able to find any existing documentation of similar issues or failures \citep[see][]{costanzachock2022audits}.
In the future, we imagine this information could become the basis for more formal consumer protections: if a user experiences harm, what are their means for recourse?
Or if they pursue legal action, how might society attribute responsibility to different entities implicated upstream?
Symmetrically, the benefits to end users do not only involve harm mitigation: for example, if a user is able to better understand how their data would be used, they might be more informed and more willing to donate their data to data collection efforts.

\paragraph{Investors.}
Much as developers benefit from \EG in that it better allows them to understand their competitors, investors benefit from \EG in that it identifies new opportunities.
For example, a venture capitalist could sort by "modality" and "created date" to understand modalities on the rise (\eg music) even before high-profile products emerge for this modality.
In turns, this suggests prime candidates for investors to take risk and fund early on the basis of concrete data and trends.
Compared to parallel public efforts in venture capital (\eg market maps documenting the foundation model ecosystem \citep{turow2023madrona} and specific startups \citep{kaufmann2022scaleindex}), \EG provides finer-grained insight by grounding to specific assets rather than institution-level trends.
We revisit the contrast between asset-centric and institution-centric analysis in \autoref{sec:related-work-dependencies}.

\paragraph{AI researchers.}
\EG provides an array of functionalities that are relevant for AI research.
However, one of the most fundamental is the increased potential for AI researchers to be aware of how foundation models are deployed.
As a concrete example, many of the applications of image-based models like Stable Diffusion diverge from what has been traditionally studied in computer vision research.
Perhaps even more clearly, billions have been invested into language-based startups whose applications (\eg copywriting) clearly differ from standard tasks studied in natural language processing (\eg natural language inference).

While academic research should not blindly follow industry trends, we do believe academic research should pay more attention to how research materializes in practice.
\EG precisely provides the opportunity for such reflection. 
Much of AI research concentrates on building better models in many senses (\eg more accurate, efficient, robust, fair):  understanding (i) what is being deployed in society, (ii) the demand for such technology, and (iii) the resulting societal impact all can help AI researchers better achieve their goals.
Similarly, many AI benchmarks are designed such that progress on the benchmark will generalize more broadly: aligning these benchmarks with deployment, or at least measuring correlations between benchmark-induced rankings and deployment-induced rankings, could prove fruitful in realizing this vision.
We specifically highlight this for work on the \textit{harms} of foundation models and AI: understanding how risks posited in the scientific literature \citep[][\textit{inter alia}]{bender2021dangers, weidinger2022taxonomy, bommasani2021opportunities, abid2021persistent, buchanan2021truth} manifest in practice could clarify how these risks should be studied, measured, mitigated, and prioritized. 

\paragraph{Economists.}
Foundation models are general-purpose technologies \citep{bresnahan1995gpt, brynjolfsson2021jcurve} that define an emerging market \citep[][\S5.5]{eloundou2023gpts, bommasani2021opportunities} worthy of study in the (digital) economy \citep{acemoglu2010skills, acemoglu2018race}.
Early work shows that foundation models can complete tasks of significant economic value \citep{noy2023experimental, felten2023chatgpt, korinek2023language}, \ie the \textit{realizable} potential of foundation models. 
\EG naturally complements this work by defining the \textit{realized} impact of foundation models at macro-scale, complementing more grounded analyses such as \citet{peng2023copilot} on developer productivity using GitHub Copilot and \citet{eloundou2023gpts} on labor exposure using GPT-4.
Put together, these works delineate inefficiencies and potential opportunities: where are foundation models not being deployed, despite showing demonstrable potential for these jobs?
More broadly, we believe \EG naturally supports efforts to understand the market structure of AI and the digital economy\citep{brynjolfsson2017ml, acemoglu2020ai, agrawal2021ai, autor2022work}.
How do pre-existing inter-organizational relationships (\eg between Microsoft and Google) interface with the new relationships mediated by the rise of foundation models?

\paragraph{Auditors.}
Auditors need to prioritize assets to allocate attention to \citep[see][]{metaxa2021audit, raji2019actionable}.
To inform these decisions, \EG provides several forms of guidance.
Most directly, auditors can prioritize (i) assets with known reports of "failures", (ii) unsatisfactory "quality control" or "analysis" information and (iii) assets with significant opacity (\ie much is unknown about the node).
Further, auditors should factor in the impact of these nodes: we recommend auditors target \textit{algorithmic monocultures} \citep{kleinberg2021monoculture, bommasani2022homogenization} that are made legible by \EG. 
Namely, if an upstream asset has extensive downstream dependencies (\eg the ChatGPT API as in \reffig{hubs}), then risks associated with this asset may propagate downstream perniciously if unchecked (\eg what if the ChatGPT API goes down?\footnote{See \url{https://status.openai.com/incidents/y6cdztrnth60}.}; what if LAION-400M is subject to data poisoning \citep[][\S4.7]{carlini2023poisoning, bommasani2021opportunities}?).
\citet{bommasani2022homogenization} directly recommend this by studying how monocultures may lead to \textit{homogeneous outcomes}: specific end users of different foundation applications being systemically failed due to the shared upstream dependence.

\paragraph{Policymakers.}
AI policy specific to foundation models is still in its infancy: \citet{bommasani2023transparency} specifically highlight the importance of transparency, directly aligning with the value proposition of \EG. 
Properties like the "access" status, "license", and "terms of service" at the scale of the ecosystem are policy-relevant: for example, policymakers should intervene if there is pervasive dependence on assets that are largely inaccessible, as these nodes will likely evade external scrutiny/oversight despite having outsized societal impact.
Beyond this, the distribution of different uses of foundation models will help clarify whether policy should adopt the perspective of regulating only the downstream applications (\eg if there is too much diversification in downstream use) or consider the upstream foundation models as well (which introduces challenges due to their generality/lack of specificity).\footnote{See the recent announcement of a UK FM taskforce: \url{https://assets.publishing.service.gov.uk/government/uploads/system/uploads/attachment_data/file/1142410/11857435_NS_IR_Refresh_2023_Supply_AllPages_Revision_4_WEB_PDF_NoCrops_-_gov.uk.pdf}.}
In testimony before the US House Subcommittee on Cybersecurity, Information Technology, and Government Innovation, \citet{madry2023supplychain} more broadly pointed to how policymakers should prioritize the emerging AI supply chain based on foundation models.
To precisely track this supply chain, and how it evolves over time, \EG provides the concrete public infrastructure.
\section{Related Work}
\label{sec:related-work}

To situate our work, we consider both our objective to make foundation models transparent and our methodology to track dependencies.

\subsection{Transparency in AI}
To make AI systems transparent, we observe three broad classes of approaches.

First, \textit{evaluation} is a widespread practice for articulating the properties and measuring the behavior of systems: in the research community, it is customary to evaluate systems against particular benchmarks to assess their performance.
Evaluations can vary in the specific type of transparency \citep[see][]{bommasani2023transparency} they provide: some evaluations quantify the accuracy of models \citep[\eg ImageNet;][]{deng2009imagenet}, others stress-test models \citep[\eg CheckList;][]{ribeiro2020beyond} or adversarially identify failures \cite[\eg red-teaming;][]{perez2022red} and still others characterize models along a range of dimensions \citep[\eg HELM;][]{liang2022helm}.
In general, while some efforts expand evaluation to datasets \citep{bommasani2020intrinsic, swayamdipta2020dataset, ethayarajh2022dataset, mitchell2022measuring} or adopt methodologies from human-computer interaction to consider human factors like user experience \citep{lee2022coauthor, lee2022interaction}, for the most part, evaluation aims to characterize a specific model in isolation.

Second, \textit{documentation} is a growing practice for specifying metadata about the broader context that situates model and system development.
Formative works like data sheets \citep{gebru2018datasheets} and model cards \citep{mitchell2018modelcards} brought this approach to the fore, complementing evaluations by articulating design decisions and developer positions involved in creating assets.
Subsequent efforts have enriched these approaches to make these documentation artifacts more useful, accessible, or otherwise aligned to specific informational needs \citep{crisan2022interactive}.\footnote{See \url{https://huggingface.co/docs/hub/model-cards}.} 
In general, documentation efforts aim to contextualize a specific asset against a broader social backdrop, often with an inclination towards how the asset came to be and with greater uptake for research-centric assets to our knowledge.

Third, \textit{analyses} and critiques have become increasingly relevant, showcasing much of the latent and oft-overlooked underpinnings of AI development and deployment.
These works often bring questions of values and power to the fore, frequently appealing to concepts or methodologies from the social sciences and disciplines beyond computer science \citep[\eg][]{dotan2020value,
ethayarajh2020utility, 
scheuerman2021politics,
raji2021ai,
koch2021reduced,
denton2021genealogy,
paullada2021data,
birhane2022values,
bommasani2022evaluation}. 
Rather than specific assets (other than for case study/illustration), analytic work centers broader themes \citep[\eg algorithmic intermediaries;][]{lazar2023algorithmiccity} or classes of technology \citep[\eg predictive optimization;][]{wang2022against}.

Our framework shares the objective of making AI (specifically foundation models) transparent. 
However, it differs along important axes from all of these established approaches, perhaps most closely resembling the documentation class (since, indeed, \EG is a documentation framework).
In contrast to current interpretations of evaluation and documentation, \EG is fundamentally about the ecosystem rather any specific asset: the value of \EG arises from tracking all assets.\footnote{We do note other works exist at ecosystem scale in other senses such AI100 \citep{stone2022artificial}, AI Index \citep{zhang2021ai}, and various data/model repositories \citep{wolf2020transformers, lhoest2021datasets}; see \url{https://modelzoo.co/}.} 
This introduces a variety of new challenges (\eg partial observability of certain information, more complicated maintenance as there is constant change across the ecosystem even if particular assets may not change for extended periods).
Further, \EG especially highlights the importance of grounding out into applications (for which a general-purpose analogue of data sheets and model cards does not exist to our knowledge) and, more generally, moving beyond research artifacts to commercial artifacts that affect broader society. 
Finally, in comparison to analytic/critical methods, \EG retains much of the concreteness of evaluation/documentation: we believe \EG provides valuable descriptive understanding that could support future normative analyses and critiques. 

Beyond these distinctions, we emphasize that our contribution extends beyond most prior works on documentation in AI.
Concretely, most prior works \citep[\eg][]{gebru2018datasheets, bender2018data, mitchell2018modelcards} propose the framework to document artifacts, perhaps with an additional proposal of who will conduct this documentation and how/why.
In contrast, we concretely execute, implementing the \EG framework in our codebase and public website. 
This mirrors works like HELM \citep{liang2022helm} where, in addition to designing an evaluation, the contributions include evaluating all language models available at present. 
The infrastructure, sustained upkeep and, ultimately, the resource itself are what provide value: ensuring transparency requires we follow through and enact the conceptual frameworks we design. 

\subsection{Dependencies}
\label{sec:related-work-dependencies}
At the technical level, \EG foreground the tracking of dependencies, whereas at the social level, \EG delineates institutional relationships.
Both of these constructs are encountered in almost every mature industry and, therefore, have been studied across a range of fields.
Concretely, almost every commercial product is the composite of some collection of materials/ingredients, meaning it has a complex supply chain.
As a result, we specifically point to related work in open-source software (which share similar implementation to \EG) and market structure (which emulate \EG in terms of organizations).

\paragraph{Open-source software.}
Much like foundation models, open-source software development is sustained by an immense network of dependencies.
Akin to our efforts to track the foundation model ecosystem, the demand to track the open-source software ecosystem is immense: the software bill of materials (SBOM) is a national-level initiative of the US's Cybersecurity and Infrastructure Security Agency to maintain an inventory of the ingredients that make up software \citep{whitehouse2021cybersecurity}.\footnote{See \url{https://www.cisa.gov/sbom}.}
These approaches have clarified how to ensure compliance from different stakeholders (\eg software vendors) and how to standardize information to support a range of use cases, providing inspiration for the abstractions we make in \EG. 
To implement this vision, a range of efforts have been put forth over the years\footnote{See \url{https://libraries.io/}, \url{http://deps.dev/}, and \url{https://ecosyste.ms/}.} with applied policy work mapping out the sociotechnical challenges for maintaining and funding these efforts \citep{ramaswami2021securing, scott2023atlanticcouncil}.
Further, they present an exemplar of policy uptake towards mandatory public reporting of these dependencies as exemplified by the proposed Securing Open Source Software Act of 2022.\footnote{\url{https://www.congress.gov/bill/117th-congress/senate-bill/4913}}
And, much akin to the \refsec{uses} we consider, these efforts already have shown how descriptive understanding of the ecosystem directly informs decision-making and characterizes the impact of assets.\footnote{See \url{https://docs.libraries.io/overview.html\#sourcerank} and \url{https://github.com/ossf/criticality\_score}.}

\paragraph{Market structure.}
In defining \EG, we made the fundamental decision to define the graph in terms of assets.
We contrast this with approaches more common in disciplines like economics and sociology, where it would be customary to instead foreground the organizations/institutions responsible for creating these assets \citep{rowlinson1997organisations}.
We believe this (fairly techno-centric) choice provides valuable leverage given the status quo: the number of assets is currently still manageable (on the order of hundreds to thousands), the assets themselves are distinctive (\eg they are not exchangeable in the way oil or steel may be in other market analyses), and specific assets markedly contextualize our understanding of organizations (\eg Stable Diffusion dramatically shapes our perception of Stability AI). 
In spite of these advantages, we point to a range of works that foreground institutions in mapping out market structure and the dynamics by which actors interact to shape the economy.
For example, given we draw upon a comparative analysis of the FM ecosystem to the automotive ecosystem, \citet{weingast1988industrial} demonstrate that institution-centrism better allow for comparisons/juxtapositions across sectors.
Alternatively, \citet{einav2010empirical} showcase how grounding to institutions facilitates various forms of measurement (\eg due to firm-level requirements on information disclosure).
Finally, many works in political and institutional sociology prime us to view institutions as the natural unit for studying power relations in modern networks and markets \cite[][\textit{inter alia}]{frickel2006new, dequech2006institutions, fleury2014sociology}.
\section{Conclusion}
\label{sec:conclusion}

The footprint of foundation models is rapidly expanding: this class of emerging technology is pervading society and is only in its infancy.
\EG aims to ensure that foundation models are transparent for a range of informational needs, thereby establishing the basic facts of how they come to impact society.
We look forward to seeing the community build upon and make use of \EG.
Concurrently, we encourage policy uptake to standardize reporting on all assets in \EG as the basis for driving change in the foundation model ecosystem.   
\begin{ack}
We thank Alex Tamkin, Ali Alvi, Amelia Hardy, Ansh Khurana, Ashwin Paranjape, Ayush Kanodia, Chris Manning, Dan Ho, Dan Jurafsky, Daniel Zhang, David Hall, Dean Carignan, Deep Ganguli, Dimitris Tsipras, Erik Brynjolfsson, Iason Gabriel, Irwing Kwong, Jack Clark, Jeremy Kaufmann, Laura Weidinger, Maggie Basta, Michael Zeng, Nazneen Rajani, Rob Reich, Rohan Taori, Tianyi Zhang, Vanessa Parli, Xia Song, Yann Dubois, and Yifan Mai for valuable feedback on this effort at various stages.
We specifically thank Ashwin Ramaswami for extensive guidance on prior work in the open-source ecosystem and future directions for the foundation model ecosystem.   
In addition, the authors would like to thank the Stanford Center for Research on Foundation Models (CRFM) and the Stanford Institute for Human-Centered Artificial Intelligence (HAI) for directly supporting this research. 
The initial release of \EG will be accompanied by a policy brief through collaboration with Stanford HAI.
RB was supported by an NSF Graduate Research Fellowship (grant number: DGE-1656518). 
\end{ack}

\bibliographystyle{unsrtnat}
\bibliography{all, main}

\begin{thebibliography}{106}
\providecommand{\natexlab}[1]{#1}
\providecommand{\url}[1]{\texttt{#1}}
\expandafter\ifx\csname urlstyle\endcsname\relax
  \providecommand{\doi}[1]{doi: #1}\else
  \providecommand{\doi}{doi: \begingroup \urlstyle{rm}\Url}\fi

\bibitem[Bommasani et~al.(2021)Bommasani, Hudson, Adeli, Altman, Arora, von
  Arx, Bernstein, Bohg, Bosselut, Brunskill, Brynjolfsson, Buch, Card,
  Castellon, Chatterji, Chen, Creel, Davis, Demszky, Donahue, Doumbouya,
  Durmus, Ermon, Etchemendy, Ethayarajh, Fei-Fei, Finn, Gale, Gillespie, Goel,
  Goodman, Grossman, Guha, Hashimoto, Henderson, Hewitt, Ho, Hong, Hsu, Huang,
  Icard, Jain, Jurafsky, Kalluri, Karamcheti, Keeling, Khani, Khattab, Koh,
  Krass, Krishna, Kuditipudi, Kumar, Ladhak, Lee, Lee, Leskovec, Levent, Li,
  Li, Ma, Malik, Manning, Mirchandani, Mitchell, Munyikwa, Nair, Narayan,
  Narayanan, Newman, Nie, Niebles, Nilforoshan, Nyarko, Ogut, Orr,
  Papadimitriou, Park, Piech, Portelance, Potts, Raghunathan, Reich, Ren, Rong,
  Roohani, Ruiz, Ryan, Ré, Sadigh, Sagawa, Santhanam, Shih, Srinivasan,
  Tamkin, Taori, Thomas, Tramèr, Wang, Wang, Wu, Wu, Wu, Xie, Yasunaga, You,
  Zaharia, Zhang, Zhang, Zhang, Zhang, Zheng, Zhou, and
  Liang]{bommasani2021opportunities}
Rishi Bommasani, Drew~A. Hudson, Ehsan Adeli, Russ Altman, Simran Arora, Sydney
  von Arx, Michael~S. Bernstein, Jeannette Bohg, Antoine Bosselut, Emma
  Brunskill, Erik Brynjolfsson, Shyamal Buch, Dallas Card, Rodrigo Castellon,
  Niladri Chatterji, Annie Chen, Kathleen Creel, Jared~Quincy Davis, Dorottya
  Demszky, Chris Donahue, Moussa Doumbouya, Esin Durmus, Stefano Ermon, John
  Etchemendy, Kawin Ethayarajh, Li~Fei-Fei, Chelsea Finn, Trevor Gale, Lauren
  Gillespie, Karan Goel, Noah Goodman, Shelby Grossman, Neel Guha, Tatsunori
  Hashimoto, Peter Henderson, John Hewitt, Daniel~E. Ho, Jenny Hong, Kyle Hsu,
  Jing Huang, Thomas Icard, Saahil Jain, Dan Jurafsky, Pratyusha Kalluri,
  Siddharth Karamcheti, Geoff Keeling, Fereshte Khani, Omar Khattab, Pang~Wei
  Koh, Mark Krass, Ranjay Krishna, Rohith Kuditipudi, Ananya Kumar, Faisal
  Ladhak, Mina Lee, Tony Lee, Jure Leskovec, Isabelle Levent, Xiang~Lisa Li,
  Xuechen Li, Tengyu Ma, Ali Malik, Christopher~D. Manning, Suvir Mirchandani,
  Eric Mitchell, Zanele Munyikwa, Suraj Nair, Avanika Narayan, Deepak
  Narayanan, Ben Newman, Allen Nie, Juan~Carlos Niebles, Hamed Nilforoshan,
  Julian Nyarko, Giray Ogut, Laurel Orr, Isabel Papadimitriou, Joon~Sung Park,
  Chris Piech, Eva Portelance, Christopher Potts, Aditi Raghunathan, Rob Reich,
  Hongyu Ren, Frieda Rong, Yusuf Roohani, Camilo Ruiz, Jack Ryan, Christopher
  Ré, Dorsa Sadigh, Shiori Sagawa, Keshav Santhanam, Andy Shih, Krishnan
  Srinivasan, Alex Tamkin, Rohan Taori, Armin~W. Thomas, Florian Tramèr,
  Rose~E. Wang, William Wang, Bohan Wu, Jiajun Wu, Yuhuai Wu, Sang~Michael Xie,
  Michihiro Yasunaga, Jiaxuan You, Matei Zaharia, Michael Zhang, Tianyi Zhang,
  Xikun Zhang, Yuhui Zhang, Lucia Zheng, Kaitlyn Zhou, and Percy Liang.
\newblock On the opportunities and risks of foundation models.
\newblock \emph{arXiv preprint arXiv:2108.07258}, 2021.

\bibitem[Devlin et~al.(2019)Devlin, Chang, Lee, and Toutanova]{devlin2019bert}
Jacob Devlin, Ming-Wei Chang, Kenton Lee, and Kristina Toutanova.
\newblock {BERT}: Pre-training of deep bidirectional transformers for language
  understanding.
\newblock In \emph{Association for Computational Linguistics (ACL)}, pages
  4171--4186, 2019.

\bibitem[Brown et~al.(2020)Brown, Mann, Ryder, Subbiah, Kaplan, Dhariwal,
  Neelakantan, Shyam, Sastry, Askell, Agarwal, Herbert-Voss, Krueger, Henighan,
  Child, Ramesh, Ziegler, Wu, Winter, Hesse, Chen, Sigler, Litwin, Gray, Chess,
  Clark, Berner, McCandlish, Radford, Sutskever, and Amodei]{brown2020gpt3}
Tom~B. Brown, Benjamin Mann, Nick Ryder, Melanie Subbiah, Jared Kaplan,
  Prafulla Dhariwal, Arvind Neelakantan, Pranav Shyam, Girish Sastry, Amanda
  Askell, Sandhini Agarwal, Ariel Herbert-Voss, Gretchen Krueger, Tom Henighan,
  Rewon Child, Aditya Ramesh, Daniel~M. Ziegler, Jeffrey Wu, Clemens Winter,
  Christopher Hesse, Mark Chen, Eric Sigler, Mateusz Litwin, Scott Gray,
  Benjamin Chess, Jack Clark, Christopher Berner, Sam McCandlish, Alec Radford,
  Ilya Sutskever, and Dario Amodei.
\newblock Language models are few-shot learners.
\newblock \emph{arXiv preprint arXiv:2005.14165}, 2020.

\bibitem[Chowdhery et~al.(2022)Chowdhery, Narang, Devlin, Bosma, Mishra,
  Roberts, Barham, Chung, Sutton, Gehrmann, Schuh, Shi, Tsvyashchenko, Maynez,
  Rao, Barnes, Tay, Shazeer, Prabhakaran, Reif, Du, Hutchinson, Pope, Bradbury,
  Austin, Isard, Gur-Ari, Yin, Duke, Levskaya, Ghemawat, Dev, Michalewski,
  García, Misra, Robinson, Fedus, Zhou, Ippolito, Luan, Lim, Zoph, Spiridonov,
  Sepassi, Dohan, Agrawal, Omernick, Dai, Pillai, Pellat, Lewkowycz, Moreira,
  Child, Polozov, Lee, Zhou, Wang, Saeta, Diaz, Firat, Catasta, Wei,
  Meier-Hellstern, Eck, Dean, Petrov, and Fiedel]{chowdhery2022palm}
Aakanksha Chowdhery, Sharan Narang, Jacob Devlin, Maarten Bosma, Gaurav Mishra,
  Adam Roberts, Paul Barham, Hyung~Won Chung, Charles Sutton, Sebastian
  Gehrmann, Parker Schuh, Kensen Shi, Sasha Tsvyashchenko, Joshua Maynez,
  A.~Rao, Parker Barnes, Yi~Tay, Noam~M. Shazeer, Vinodkumar Prabhakaran, Emily
  Reif, Nan Du, B.~Hutchinson, Reiner Pope, James Bradbury, Jacob Austin,
  M.~Isard, Guy Gur-Ari, Pengcheng Yin, Toju Duke, Anselm Levskaya,
  S.~Ghemawat, Sunipa Dev, Henryk Michalewski, Xavier García, Vedant Misra,
  Kevin Robinson, Liam Fedus, Denny Zhou, Daphne Ippolito, D.~Luan, Hyeontaek
  Lim, Barret Zoph, A.~Spiridonov, Ryan Sepassi, David Dohan, Shivani Agrawal,
  Mark Omernick, Andrew~M. Dai, T.~S. Pillai, Marie Pellat, Aitor Lewkowycz,
  E.~Moreira, Rewon Child, Oleksandr Polozov, Katherine Lee, Zongwei Zhou,
  Xuezhi Wang, Brennan Saeta, Mark Diaz, Orhan Firat, Michele Catasta, Jason
  Wei, K.~Meier-Hellstern, D.~Eck, J.~Dean, Slav Petrov, and Noah Fiedel.
\newblock {PaLM}: Scaling language modeling with pathways.
\newblock \emph{arXiv}, 2022.

\bibitem[Chen et~al.(2020)Chen, Radford, Child, Wu, Jun, Luan, and
  Sutskever]{chen2020imagegpt}
Mark Chen, Alec Radford, Rewon Child, Jeffrey Wu, Heewoo Jun, David Luan, and
  Ilya Sutskever.
\newblock Generative pretraining from pixels.
\newblock In Hal~Daumé III and Aarti Singh, editors, \emph{Proceedings of the
  37th International Conference on Machine Learning}, volume 119 of
  \emph{Proceedings of Machine Learning Research}, pages 1691--1703. PMLR,
  13--18 Jul 2020.
\newblock URL \url{http://proceedings.mlr.press/v119/chen20s.html}.

\bibitem[Ramesh et~al.(2021)Ramesh, Pavlov, Goh, Gray, Voss, Radford, Chen, and
  Sutskever]{ramesh2021dalle}
Aditya Ramesh, Mikhail Pavlov, Gabriel Goh, Scott Gray, Chelsea Voss, Alec
  Radford, Mark Chen, and Ilya Sutskever.
\newblock Zero-shot text-to-image generation.
\newblock In Marina Meila and Tong Zhang, editors, \emph{Proceedings of the
  38th International Conference on Machine Learning}, volume 139 of
  \emph{Proceedings of Machine Learning Research}, pages 8821--8831. PMLR,
  18--24 Jul 2021.
\newblock URL \url{https://proceedings.mlr.press/v139/ramesh21a.html}.

\bibitem[Radford et~al.(2021)Radford, Kim, Hallacy, Ramesh, Goh, Agarwal,
  Sastry, Askell, Mishkin, Clark, Krueger, and Sutskever]{radford2021clip}
Alec Radford, Jong~Wook Kim, Chris Hallacy, Aditya Ramesh, Gabriel Goh,
  Sandhini Agarwal, Girish Sastry, Amanda Askell, Pamela Mishkin, Jack Clark,
  Gretchen Krueger, and Ilya Sutskever.
\newblock Learning transferable visual models from natural language
  supervision.
\newblock In \emph{International Conference on Machine Learning (ICML)}, volume
  139, pages 8748--8763, 2021.

\bibitem[Singer et~al.(2022)Singer, Polyak, Hayes, Yin, An, Zhang, Hu, Yang,
  Ashual, Gafni, Parikh, Gupta, and Taigman]{singer2022makeavideo}
Uriel Singer, Adam Polyak, Thomas Hayes, Xiaoyue Yin, Jie An, Songyang Zhang,
  Qiyuan Hu, Harry Yang, Oron Ashual, Oran Gafni, Devi Parikh, Sonal Gupta, and
  Yaniv Taigman.
\newblock Make-a-video: Text-to-video generation without text-video data.
\newblock \emph{ArXiv}, abs/2209.14792, 2022.

\bibitem[Wang et~al.(2022{\natexlab{a}})Wang, Li, Li, He, Huang, Zhao, Zhang,
  Xu, Liu, Wang, Xing, Chen, Pan, Yu, Wang, Wang, and
  Qiao]{wang2022internvideo}
Yi~Wang, Kunchang Li, Yizhuo Li, Yinan He, Bingkun Huang, Zhiyu Zhao, Hongjie
  Zhang, Jilan Xu, Yi~Liu, Zun Wang, Sen Xing, Guo Chen, Junting Pan, Jiashuo
  Yu, Yali Wang, Limin Wang, and Yu~Qiao.
\newblock Internvideo: General video foundation models via generative and
  discriminative learning.
\newblock \emph{ArXiv}, abs/2212.03191, 2022{\natexlab{a}}.

\bibitem[Chen et~al.(2021)Chen, Tworek, Jun, Yuan, Ponde, Kaplan, Edwards,
  Burda, Joseph, Brockman, Ray, Puri, Krueger, Petrov, Khlaaf, Sastry, Mishkin,
  Chan, Gray, Ryder, Pavlov, Power, Kaiser, Bavarian, Winter, Tillet, Such,
  Cummings, Plappert, Chantzis, Barnes, Herbert-Voss, Guss, Nichol, Babuschkin,
  Balaji, Jain, Carr, Leike, Achiam, Misra, Morikawa, Radford, Knight,
  Brundage, Murati, Mayer, Welinder, McGrew, Amodei, McCandlish, Sutskever, and
  Zaremba]{chen2021codex}
Mark Chen, Jerry Tworek, Heewoo Jun, Qiming Yuan, Henrique Ponde, Jared Kaplan,
  Harrison Edwards, Yura Burda, Nicholas Joseph, Greg Brockman, Alex Ray, Raul
  Puri, Gretchen Krueger, Michael Petrov, Heidy Khlaaf, Girish Sastry, Pamela
  Mishkin, Brooke Chan, Scott Gray, Nick Ryder, Mikhail Pavlov, Alethea Power,
  Lukasz Kaiser, Mohammad Bavarian, Clemens Winter, Philippe Tillet,
  Felipe~Petroski Such, David~W. Cummings, Matthias Plappert, Fotios Chantzis,
  Elizabeth Barnes, Ariel Herbert-Voss, William~H. Guss, Alex Nichol, Igor
  Babuschkin, S.~Arun Balaji, Shantanu Jain, Andrew Carr, Jan Leike, Joshua
  Achiam, Vedant Misra, Evan Morikawa, Alec Radford, Matthew~M. Knight, Miles
  Brundage, Mira Murati, Katie Mayer, Peter Welinder, Bob McGrew, Dario Amodei,
  Sam McCandlish, Ilya Sutskever, and Wojciech Zaremba.
\newblock Evaluating large language models trained on code.
\newblock \emph{ArXiv}, abs/2107.03374, 2021.

\bibitem[Jumper et~al.(2021)Jumper, Evans, Pritzel, Green, Figurnov,
  Ronneberger, Tunyasuvunakool, Bates, Z{\'i}dek, Potapenko, Bridgland, Meyer,
  Kohl, Ballard, Cowie, Romera-Paredes, Nikolov, Jain, Adler, Back, Petersen,
  Reiman, Clancy, Zielinski, Steinegger, Pacholska, Berghammer, Bodenstein,
  Silver, Vinyals, Senior, Kavukcuoglu, Kohli, and
  Hassabis]{jumper2021alphafold}
John~M. Jumper, Richard Evans, Alexander Pritzel, Tim Green, Michael Figurnov,
  Olaf Ronneberger, Kathryn Tunyasuvunakool, Russ Bates, Augustin Z{\'i}dek,
  Anna Potapenko, Alex Bridgland, Clemens Meyer, Simon A~A Kohl, Andy Ballard,
  Andrew Cowie, Bernardino Romera-Paredes, Stanislav Nikolov, Rishub Jain,
  Jonas Adler, Trevor Back, Stig Petersen, David~A. Reiman, Ellen Clancy,
  Michal Zielinski, Martin Steinegger, Michalina Pacholska, Tamas Berghammer,
  Sebastian Bodenstein, David Silver, Oriol Vinyals, Andrew~W. Senior, Koray
  Kavukcuoglu, Pushmeet Kohli, and Demis Hassabis.
\newblock Highly accurate protein structure prediction with alphafold.
\newblock \emph{Nature}, 596:\penalty0 583 -- 589, 2021.

\bibitem[Verkuil et~al.(2022)Verkuil, Kabeli, Du, Wicky, Milles, Dauparas,
  Baker, Ovchinnikov, Sercu, and Rives]{verkuil2022esm2}
Robert Verkuil, Ori Kabeli, Yilun Du, Basile I.~M. Wicky, Lukas~F. Milles,
  Justas Dauparas, David Baker, Sergey Ovchinnikov, Tom Sercu, and Alexander
  Rives.
\newblock Language models generalize beyond natural proteins.
\newblock \emph{bioRxiv}, 2022.
\newblock \doi{10.1101/2022.12.21.521521}.
\newblock URL
  \url{https://www.biorxiv.org/content/early/2022/12/22/2022.12.21.521521}.

\bibitem[NYT(2020)]{nyt2020}
NYT.
\newblock Meet gpt-3. it has learned to code (and blog and argue).
\newblock 2020.
\newblock URL
  \url{https://www.nytimes.com/2020/11/24/science/artificial-intelligence-ai-gpt3.html}.

\bibitem[Nature(2021)]{nature2021}
Nature.
\newblock The big question.
\newblock 2021.
\newblock URL \url{https://www.nature.com/articles/s42256-021-00395-y}.

\bibitem[Economist(2022)]{economist2022}
Economist.
\newblock Huge “foundation models” are turbo-charging ai progress.
\newblock 2022.
\newblock URL
  \url{https://www.economist.com/interactive/briefing/2022/06/11/huge-foundation-models-are-turbo-charging-ai-progress}.

\bibitem[CNN(2023)]{cnn2023}
CNN.
\newblock Why you’re about to see chatgpt in more of your apps.
\newblock 2023.
\newblock URL \url{https://www.cnn.com/2023/03/01/tech/chatgpt-api/index.html}.

\bibitem[Hu(2023)]{hu2023chatgpt}
Krystal Hu.
\newblock Chatgpt sets record for fastest-growing user base - analyst note.
\newblock 2023.
\newblock URL
  \url{https://www.reuters.com/technology/chatgpt-sets-record-fastest-growing-user-base-analyst-note-2023-02-01/}.

\bibitem[Appenzeller et~al.(2022)Appenzeller, Bornstein, Casado, and
  Li]{appenzeller2022stablediffusion}
Guido Appenzeller, Matt Bornstein, Martin Casado, and Yoko Li.
\newblock Art isn’t dead, it’s just machine-generated.
\newblock 2022.
\newblock URL \url{https://a16z.com/2022/11/16/creativity-as-an-app/}.

\bibitem[Kaufmann et~al.(2022)Kaufmann, Abram, and
  Basta]{kaufmann2022scaleindex}
Jeremy Kaufmann, Max Abram, and Maggie Basta.
\newblock Introducing: the scale generative ai index.
\newblock 2022.
\newblock URL
  \url{https://www.scalevp.com/blog/introducing-the-scale-generative-ai-index}.

\bibitem[Wei et~al.(2022)Wei, Tay, Bommasani, Raffel, Zoph, Borgeaud, Yogatama,
  Bosma, Zhou, Metzler, Chi, Hashimoto, Vinyals, Liang, Dean, and
  Fedus]{wei2022emergent}
Jason Wei, Yi~Tay, Rishi Bommasani, Colin Raffel, Barret Zoph, Sebastian
  Borgeaud, Dani Yogatama, Maarten Bosma, Denny Zhou, Donald Metzler, Ed~H.
  Chi, Tatsunori Hashimoto, Oriol Vinyals, Percy Liang, Jeff Dean, and William
  Fedus.
\newblock Emergent abilities of large language models.
\newblock \emph{Transactions on Machine Learning Research}, 2022.
\newblock URL \url{https://openreview.net/forum?id=yzkSU5zdwD}.
\newblock Survey Certification.

\bibitem[Abid et~al.(2021)Abid, Farooqi, and Zou]{abid2021persistent}
Abubakar Abid, Maheen Farooqi, and James Zou.
\newblock Persistent anti-muslim bias in large language models.
\newblock \emph{arXiv preprint arXiv:2101.05783}, 2021.

\bibitem[Gao et~al.(2021)Gao, Biderman, Black, Golding, Hoppe, Foster, Phang,
  He, Thite, Nabeshima, Presser, and Leahy]{gao2021thepile}
Leo Gao, Stella Biderman, Sid Black, Laurence Golding, Travis Hoppe, Charles
  Foster, Jason Phang, Horace He, Anish Thite, Noa Nabeshima, Shawn Presser,
  and Connor Leahy.
\newblock {T}he {P}ile: {A}n 800{GB} {D}ataset of {D}iverse {T}ext for
  {L}anguage {M}odeling.
\newblock \emph{arXiv preprint arXiv:2101.00027}, 2021.
\newblock URL \url{https://arxiv.org/abs/2101.00027}.

\bibitem[Sanh et~al.(2021)Sanh, Webson, Raffel, Bach, Sutawika, Alyafeai,
  Chaffin, Stiegler, Scao, Raja, Dey, Bari, Xu, Thakker, Sharma, Szczechla,
  Kim, Chhablani, Nayak, Datta, Chang, Jiang, Wang, Manica, Shen, Yong, Pandey,
  Bawden, Wang, Neeraj, Rozen, Sharma, Santilli, Fevry, Fries, Teehan,
  Biderman, Gao, Bers, Wolf, and Rush]{sanh2021multitask}
Victor Sanh, Albert Webson, Colin Raffel, Stephen~H. Bach, Lintang Sutawika,
  Zaid Alyafeai, Antoine Chaffin, Arnaud Stiegler, Teven~Le Scao, Arun Raja,
  Manan Dey, M~Saiful Bari, Canwen Xu, Urmish Thakker, Shanya~Sharma Sharma,
  Eliza Szczechla, Taewoon Kim, Gunjan Chhablani, Nihal Nayak, Debajyoti Datta,
  Jonathan Chang, Mike Tian-Jian Jiang, Han Wang, Matteo Manica, Sheng Shen,
  Zheng~Xin Yong, Harshit Pandey, Rachel Bawden, Thomas Wang, Trishala Neeraj,
  Jos Rozen, Abheesht Sharma, Andrea Santilli, Thibault Fevry, Jason~Alan
  Fries, Ryan Teehan, Stella Biderman, Leo Gao, Tali Bers, Thomas Wolf, and
  Alexander~M. Rush.
\newblock Multitask prompted training enables zero-shot task generalization.
\newblock \emph{arXiv}, 2021.

\bibitem[OpenAI(2023{\natexlab{a}})]{openai2023chatgptapi}
OpenAI.
\newblock Introducing chatgpt and whisper apis.
\newblock 2023{\natexlab{a}}.
\newblock URL
  \url{https://openai.com/blog/introducing-chatgpt-and-whisper-apis}.

\bibitem[Gebru et~al.(2018)Gebru, Morgenstern, Vecchione, Vaughan, Wallach,
  Ill, and Crawford]{gebru2018datasheets}
Timnit Gebru, Jamie Morgenstern, Briana Vecchione, Jennifer~Wortman Vaughan,
  Hanna Wallach, Hal~Daumé Ill, and Kate Crawford.
\newblock Datasheets for datasets.
\newblock \emph{arXiv preprint arXiv:1803.09010}, 2018.

\bibitem[Bender and Friedman(2018)]{bender2018data}
Emily~M Bender and Batya Friedman.
\newblock Data statements for natural language processing: Toward mitigating
  system bias and enabling better science.
\newblock \emph{Transactions of the Association for Computational Linguistics
  (TACL)}, 6:\penalty0 587--604, 2018.

\bibitem[Mitchell et~al.(2018)Mitchell, Wu, Zaldivar, Barnes, Vasserman,
  Hutchinson, Spitzer, Raji, and Gebru]{mitchell2018modelcards}
Margaret Mitchell, Simone Wu, Andrew Zaldivar, Parker Barnes, Lucy Vasserman,
  Ben Hutchinson, Elena Spitzer, Inioluwa~Deborah Raji, and Timnit Gebru.
\newblock Model cards for model reporting.
\newblock \emph{Proceedings of the Conference on Fairness, Accountability, and
  Transparency}, 2018.

\bibitem[Kleinberg(1999)]{kleinberg1999hubs}
Jon~M. Kleinberg.
\newblock Hubs, authorities, and communities.
\newblock \emph{ACM Comput. Surv.}, 31\penalty0 (4es):\penalty0 5–es, dec
  1999.
\newblock ISSN 0360-0300.
\newblock \doi{10.1145/345966.345982}.
\newblock URL \url{https://doi.org/10.1145/345966.345982}.

\bibitem[Hendricks et~al.(1995)Hendricks, Piccione, and Tan]{hendricks1995hubs}
Ken Hendricks, Michele Piccione, and Guofu Tan.
\newblock {The Economics of Hubs: The Case of Monopoly}.
\newblock \emph{The Review of Economic Studies}, 62\penalty0 (1):\penalty0
  83--99, 01 1995.
\newblock ISSN 0034-6527.
\newblock \doi{10.2307/2297842}.
\newblock URL \url{https://doi.org/10.2307/2297842}.

\bibitem[Franks et~al.(2008)Franks, Noble, Kaufmann, and
  Stagl]{franks2008extremism}
Daniel~W Franks, Jason Noble, Peter Kaufmann, and Sigrid Stagl.
\newblock Extremism propagation in social networks with hubs.
\newblock \emph{Adaptive Behavior}, 16\penalty0 (4):\penalty0 264--274, 2008.

\bibitem[Van~den Heuvel and Sporns(2013)]{van2013network}
Martijn~P Van~den Heuvel and Olaf Sporns.
\newblock Network hubs in the human brain.
\newblock \emph{Trends in cognitive sciences}, 17\penalty0 (12):\penalty0
  683--696, 2013.

\bibitem[Martin~Jr et~al.(2020)Martin~Jr, Prabhakaran, Kuhlberg, Smart, and
  Isaac]{martin2020extending}
Donald Martin~Jr, Vinodkumar Prabhakaran, Jill Kuhlberg, Andrew Smart, and
  William~S Isaac.
\newblock Extending the machine learning abstraction boundary: A complex
  systems approach to incorporate societal context.
\newblock \emph{arXiv preprint arXiv:2006.09663}, 2020.

\bibitem[Amironesei et~al.(2021)Amironesei, Denton, and
  Hanna]{amironesei2021notes}
Razvan Amironesei, Emily Denton, and Alex Hanna.
\newblock Notes on problem formulation in machine learning.
\newblock \emph{IEEE Technology and Society Magazine}, 40\penalty0
  (3):\penalty0 80--83, 2021.
\newblock \doi{10.1109/MTS.2021.3104380}.

\bibitem[Paullada et~al.(2021)Paullada, Raji, Bender, Denton, and
  Hanna]{paullada2021data}
Amandalynne Paullada, Inioluwa~Deborah Raji, Emily~M. Bender, Emily Denton, and
  Alex Hanna.
\newblock Data and its (dis)contents: A survey of dataset development and use
  in machine learning research.
\newblock \emph{Patterns}, 2\penalty0 (11):\penalty0 100336, 2021.
\newblock ISSN 2666-3899.
\newblock \doi{https://doi.org/10.1016/j.patter.2021.100336}.
\newblock URL
  \url{https://www.sciencedirect.com/science/article/pii/S2666389921001847}.

\bibitem[Birhane et~al.(2022)Birhane, Kalluri, Card, Agnew, Dotan, and
  Bao]{birhane2022values}
Abeba Birhane, Pratyusha Kalluri, Dallas Card, William Agnew, Ravit Dotan, and
  Michelle Bao.
\newblock The values encoded in machine learning research.
\newblock In \emph{2022 ACM Conference on Fairness, Accountability, and
  Transparency}, FAccT '22, page 173–184, New York, NY, USA, 2022.
  Association for Computing Machinery.
\newblock ISBN 9781450393522.
\newblock \doi{10.1145/3531146.3533083}.
\newblock URL \url{https://doi.org/10.1145/3531146.3533083}.

\bibitem[Rombach et~al.(2021)Rombach, Blattmann, Lorenz, Esser, and
  Ommer]{rombach2021highresolution}
Robin Rombach, Andreas Blattmann, Dominik Lorenz, Patrick Esser, and Björn
  Ommer.
\newblock High-resolution image synthesis with latent diffusion models, 2021.

\bibitem[Schuhmann et~al.(2022)Schuhmann, Beaumont, Vencu, Gordon, Wightman,
  Cherti, Coombes, Katta, Mullis, Wortsman, Schramowski, Kundurthy, Crowson,
  Schmidt, Kaczmarczyk, and Jitsev]{schuhmann2022laion}
Christoph Schuhmann, Romain Beaumont, Richard Vencu, Cade~W Gordon, Ross
  Wightman, Mehdi Cherti, Theo Coombes, Aarush Katta, Clayton Mullis, Mitchell
  Wortsman, Patrick Schramowski, Srivatsa~R Kundurthy, Katherine Crowson,
  Ludwig Schmidt, Robert Kaczmarczyk, and Jenia Jitsev.
\newblock {LAION}-5b: An open large-scale dataset for training next generation
  image-text models.
\newblock In \emph{Thirty-sixth Conference on Neural Information Processing
  Systems Datasets and Benchmarks Track}, 2022.
\newblock URL \url{https://openreview.net/forum?id=M3Y74vmsMcY}.

\bibitem[OpenAI(2023{\natexlab{b}})]{openai2023gpt4}
OpenAI.
\newblock Gpt-4 technical report.
\newblock 2023{\natexlab{b}}.

\bibitem[Lieber et~al.(2021)Lieber, Sharir, Lenz, and
  Shoham]{lieber2021jurassic}
Opher Lieber, Or~Sharir, Barak Lenz, and Yoav Shoham.
\newblock Jurassic-1: Technical details and evaluation.
\newblock \emph{White Paper, AI21 Labs}, 2021.
\newblock URL
  \url{https://uploads-ssl.webflow.com/60fd4503684b466578c0d307/61138924626a6981ee09caf6_jurassic_tech_paper.pdf}.

\bibitem[Liang et~al.(2022{\natexlab{a}})Liang, Bommasani, Lee, Tsipras, Soylu,
  Yasunaga, Zhang, Narayanan, Wu, Kumar, Newman, Yuan, Yan, Zhang, Cosgrove,
  Manning, R'e, Acosta-Navas, Hudson, Zelikman, Durmus, Ladhak, Rong, Ren, Yao,
  Wang, Santhanam, Orr, Zheng, Yuksekgonul, Suzgun, Kim, Guha, Chatterji,
  Khattab, Henderson, Huang, Chi, Xie, Santurkar, Ganguli, Hashimoto, Icard,
  Zhang, Chaudhary, Wang, Li, Mai, Zhang, and Koreeda]{liang2022helm}
Percy Liang, Rishi Bommasani, Tony Lee, Dimitris Tsipras, Dilara Soylu,
  Michihiro Yasunaga, Yian Zhang, Deepak Narayanan, Yuhuai Wu, Ananya Kumar,
  Benjamin Newman, Binhang Yuan, Bobby Yan, Ce~Zhang, Christian Cosgrove,
  Christopher~D. Manning, Christopher R'e, Diana Acosta-Navas, Drew~A. Hudson,
  E.~Zelikman, Esin Durmus, Faisal Ladhak, Frieda Rong, Hongyu Ren, Huaxiu Yao,
  Jue Wang, Keshav Santhanam, Laurel~J. Orr, Lucia Zheng, Mert Yuksekgonul,
  Mirac Suzgun, Nathan~S. Kim, Neel Guha, Niladri~S. Chatterji, O.~Khattab,
  Peter Henderson, Qian Huang, Ryan Chi, Sang~Michael Xie, Shibani Santurkar,
  Surya Ganguli, Tatsunori Hashimoto, Thomas~F. Icard, Tianyi Zhang, Vishrav
  Chaudhary, William Wang, Xuechen Li, Yifan Mai, Yuhui Zhang, and Yuta
  Koreeda.
\newblock Holistic evaluation of language models.
\newblock \emph{ArXiv}, abs/2211.09110, 2022{\natexlab{a}}.

\bibitem[Dodge et~al.(2021)Dodge, Sap, Marasovi{\'c}, Agnew, Ilharco,
  Groeneveld, Mitchell, and Gardner]{dodge2021c4}
Jesse Dodge, Maarten Sap, Ana Marasovi{\'c}, William Agnew, Gabriel Ilharco,
  Dirk Groeneveld, Margaret Mitchell, and Matt Gardner.
\newblock Documenting large webtext corpora: A case study on the colossal clean
  crawled corpus.
\newblock In \emph{Proceedings of the 2021 Conference on Empirical Methods in
  Natural Language Processing}, pages 1286--1305, Online and Punta Cana,
  Dominican Republic, November 2021. Association for Computational Linguistics.
\newblock \doi{10.18653/v1/2021.emnlp-main.98}.
\newblock URL \url{https://aclanthology.org/2021.emnlp-main.98}.

\bibitem[Gururangan et~al.(2022)Gururangan, Card, Drier, Gade, Wang, Wang,
  Zettlemoyer, and Smith]{gururangan2022whose}
Suchin Gururangan, Dallas Card, Sarah~K. Drier, Emily~Kalah Gade, Leroy~Z.
  Wang, Zeyu Wang, Luke Zettlemoyer, and Noah~A. Smith.
\newblock Whose language counts as high quality? measuring language ideologies
  in text data selection.
\newblock In \emph{Conference on Empirical Methods in Natural Language
  Processing}, 2022.

\bibitem[Lacoste et~al.(2019)Lacoste, Luccioni, Schmidt, and
  Dandres]{lacoste2019quantifying}
Alexandre Lacoste, Alexandra Luccioni, Victor Schmidt, and Thomas Dandres.
\newblock Quantifying the carbon emissions of machine learning.
\newblock \emph{arXiv preprint arXiv:1910.09700}, 2019.

\bibitem[Strubell et~al.(2019)Strubell, Ganesh, and
  McCallum]{strubell2019energy}
Emma Strubell, Ananya Ganesh, and Andrew McCallum.
\newblock Energy and policy considerations for deep learning in {NLP}.
\newblock \emph{arXiv preprint arXiv:1906.02243}, 2019.

\bibitem[Henderson et~al.(2020)Henderson, Hu, Romoff, Brunskill, Jurafsky, and
  Pineau]{henderson2020towards}
Peter Henderson, Jieru Hu, Joshua Romoff, Emma Brunskill, Dan Jurafsky, and
  Joelle Pineau.
\newblock Towards the systematic reporting of the energy and carbon footprints
  of machine learning.
\newblock \emph{Journal of Machine Learning Research}, 21\penalty0
  (248):\penalty0 1--43, 2020.

\bibitem[Luccioni and Hern{\'a}ndez-Garc{\'i}a(2023)]{luccioni2023counting}
Alexandra~Sasha Luccioni and Alex Hern{\'a}ndez-Garc{\'i}a.
\newblock Counting carbon: A survey of factors influencing the emissions of
  machine learning.
\newblock \emph{ArXiv}, abs/2302.08476, 2023.

\bibitem[Carlini et~al.(2023)Carlini, Jagielski, Choquette-Choo, Paleka,
  Pearce, Anderson, Terzis, Thomas, and Tram{\`e}r]{carlini2023poisoning}
Nicholas Carlini, Matthew Jagielski, Christopher~A. Choquette-Choo, Daniel
  Paleka, Will Pearce, H.~Anderson, A.~Terzis, Kurt Thomas, and Florian
  Tram{\`e}r.
\newblock Poisoning web-scale training datasets is practical.
\newblock \emph{ArXiv}, abs/2302.10149, 2023.

\bibitem[Bommasani et~al.(2023)Bommasani, Zhang, Lee, and
  Liang]{bommasani2023transparency}
Rishi Bommasani, Daniel Zhang, Tony Lee, and Percy Liang.
\newblock Improving transparency in ai language models: A holistic evaluation.
\newblock \emph{Foundation Model Issue Brief Series}, 2023.
\newblock URL
  \url{https://hai.stanford.edu/foundation-model-issue-brief-series}.

\bibitem[Mądry(2023)]{madry2023supplychain}
Aleksander Mądry.
\newblock Advances in ai: Are we ready for a tech revolution?
\newblock \emph{Cybersecurity, Information Technology, and Government
  Innovation Subcommittee}, 2023.
\newblock URL
  \url{https://oversight.house.gov/wp-content/uploads/2023/03/madry_written_statement100.pdf}.

\bibitem[Zhao et~al.(2017)Zhao, Laszka, and Grossklags]{zhao2017devising}
Mingyi Zhao, Aron Laszka, and Jens Grossklags.
\newblock Devising effective policies for bug-bounty platforms and security
  vulnerability discovery.
\newblock \emph{Journal of Information Policy}, 7:\penalty0 372--418, 2017.

\bibitem[Chowdhury and Williams(2021)]{chowdhury2021twitter}
Rumman Chowdhury and Jutta Williams.
\newblock Introducing twitter’s first algorithmic bias bounty challenge.
\newblock 2021.
\newblock URL
  \url{https://blog.twitter.com/engineering/en_us/topics/insights/2021/algorithmic-bias-bounty-challenge}.

\bibitem[Liang et~al.(2022{\natexlab{b}})Liang, Bommasani, Creel, and
  Reich]{liang2022community-norms}
Percy Liang, Rishi Bommasani, Kathleen~A. Creel, and Rob Reich.
\newblock The time is now to develop community norms for the release of
  foundation models, 2022{\natexlab{b}}.
\newblock URL \url{https://crfm.stanford.edu/2022/05/17/community-norms.html}.

\bibitem[Costanza-Chock et~al.(2022)Costanza-Chock, Raji, and
  Buolamwini]{costanzachock2022audits}
Sasha Costanza-Chock, Inioluwa~Deborah Raji, and Joy Buolamwini.
\newblock Who audits the auditors? recommendations from a field scan of the
  algorithmic auditing ecosystem.
\newblock In \emph{2022 ACM Conference on Fairness, Accountability, and
  Transparency}, FAccT '22, page 1571–1583, New York, NY, USA, 2022.
  Association for Computing Machinery.
\newblock ISBN 9781450393522.
\newblock \doi{10.1145/3531146.3533213}.
\newblock URL \url{https://doi.org/10.1145/3531146.3533213}.

\bibitem[Turow et~al.(2023)Turow, Goel, and Porter]{turow2023madrona}
Jon Turow, Palak Goel, and Tim Porter.
\newblock Foundation models: The future (still) isn’t happening fast enough.
\newblock \emph{Madrona}, 2023.
\newblock URL \url{https://www.madrona.com/foundation-models/}.

\bibitem[Bender et~al.(2021)Bender, Gebru, McMillan-Major, and
  Shmitchell]{bender2021dangers}
Emily~M Bender, Timnit Gebru, Angelina McMillan-Major, and Shmargaret
  Shmitchell.
\newblock On the dangers of stochastic parrots: Can language models be too big?
\newblock In \emph{Proceedings of the 2021 ACM Conference on Fairness,
  Accountability, and Transparency}, pages 610--623, 2021.

\bibitem[Weidinger et~al.(2022)Weidinger, Uesato, Rauh, Griffin, Huang, Mellor,
  Glaese, Cheng, Balle, Kasirzadeh, Biles, Brown, Kenton, Hawkins, Stepleton,
  Birhane, Hendricks, Rimell, Isaac, Haas, Legassick, Irving, and
  Gabriel]{weidinger2022taxonomy}
Laura Weidinger, Jonathan Uesato, Maribeth Rauh, Conor Griffin, Po-Sen Huang,
  John Mellor, Amelia Glaese, Myra Cheng, Borja Balle, Atoosa Kasirzadeh,
  Courtney Biles, Sasha Brown, Zac Kenton, Will Hawkins, Tom Stepleton, Abeba
  Birhane, Lisa~Anne Hendricks, Laura Rimell, William Isaac, Julia Haas, Sean
  Legassick, Geoffrey Irving, and Iason Gabriel.
\newblock Taxonomy of risks posed by language models.
\newblock In \emph{2022 ACM Conference on Fairness, Accountability, and
  Transparency}, FAccT '22, page 214–229, New York, NY, USA, 2022.
  Association for Computing Machinery.
\newblock ISBN 9781450393522.
\newblock \doi{10.1145/3531146.3533088}.
\newblock URL \url{https://doi.org/10.1145/3531146.3533088}.

\bibitem[Buchanan et~al.(2021)Buchanan, Lohn, Musser, and
  Sedova]{buchanan2021truth}
Ben Buchanan, Andrew Lohn, Micah Musser, and Katerina Sedova.
\newblock Truth, lies, and automation.
\newblock 2021.

\bibitem[Bresnahan and Trajtenberg(1995)]{bresnahan1995gpt}
Timothy~F. Bresnahan and M.~Trajtenberg.
\newblock General purpose technologies ‘engines of growth’?
\newblock \emph{Journal of Econometrics}, 65\penalty0 (1):\penalty0 83--108,
  1995.
\newblock ISSN 0304-4076.
\newblock \doi{https://doi.org/10.1016/0304-4076(94)01598-T}.
\newblock URL
  \url{https://www.sciencedirect.com/science/article/pii/030440769401598T}.

\bibitem[Brynjolfsson et~al.(2021)Brynjolfsson, Rock, and
  Syverson]{brynjolfsson2021jcurve}
Erik Brynjolfsson, Daniel Rock, and Chad Syverson.
\newblock The productivity j-curve: How intangibles complement general purpose
  technologies.
\newblock \emph{American Economic Journal: Macroeconomics}, 13\penalty0
  (1):\penalty0 333--72, January 2021.
\newblock \doi{10.1257/mac.20180386}.
\newblock URL \url{https://www.aeaweb.org/articles?id=10.1257/mac.20180386}.

\bibitem[Eloundou et~al.(2023)Eloundou, Manning, Mishkin, and
  Rock]{eloundou2023gpts}
Tyna Eloundou, Sam Manning, Pamela Mishkin, and Daniel Rock.
\newblock Gpts are gpts: An early look at the labor market impact potential of
  large language models, 2023.

\bibitem[Acemoglu and Autor(2010)]{acemoglu2010skills}
Daron Acemoglu and David Autor.
\newblock Skills, tasks and technologies: Implications for employment and
  earnings.
\newblock Working Paper 16082, National Bureau of Economic Research, June 2010.
\newblock URL \url{http://www.nber.org/papers/w16082}.

\bibitem[Acemoglu and Restrepo(2018)]{acemoglu2018race}
Daron Acemoglu and Pascual Restrepo.
\newblock The race between man and machine: Implications of technology for
  growth, factor shares, and employment.
\newblock \emph{American Economic Review}, 108\penalty0 (6):\penalty0
  1488--1542, June 2018.
\newblock \doi{10.1257/aer.20160696}.
\newblock URL \url{https://www.aeaweb.org/articles?id=10.1257/aer.20160696}.

\bibitem[Noy and Zhang(2023)]{noy2023experimental}
Shakked Noy and Whitney Zhang.
\newblock Experimental evidence on the productivity effects of generative
  artificial intelligence.
\newblock \emph{SSRN Electronic Journal}, 2023.

\bibitem[Felten et~al.(2023)Felten, Raj, and Seamans]{felten2023chatgpt}
Edward~W. Felten, Manav Raj, and Robert~C. Seamans.
\newblock How will language modelers like chatgpt affect occupations and
  industries?
\newblock \emph{SSRN Electronic Journal}, 2023.

\bibitem[Korinek(2023)]{korinek2023language}
Anton Korinek.
\newblock Language models and cognitive automation for economic research.
\newblock NBER Working Papers 30957, National Bureau of Economic Research, Inc,
  2023.
\newblock URL \url{https://EconPapers.repec.org/RePEc:nbr:nberwo:30957}.

\bibitem[Peng et~al.(2023)Peng, Kalliamvakou, Cihon, and
  Demirer]{peng2023copilot}
Sida Peng, Eirini Kalliamvakou, Peter Cihon, and Mert Demirer.
\newblock The impact of ai on developer productivity: Evidence from github
  copilot.
\newblock \emph{ArXiv}, abs/2302.06590, 2023.

\bibitem[Brynjolfsson and Mitchell(2017)]{brynjolfsson2017ml}
Erik Brynjolfsson and Tom Mitchell.
\newblock What can machine learning do? workforce implications.
\newblock \emph{Science}, 358\penalty0 (6370):\penalty0 1530--1534, 2017.
\newblock \doi{10.1126/science.aap8062}.
\newblock URL \url{https://www.science.org/doi/abs/10.1126/science.aap8062}.

\bibitem[Acemoglu et~al.(2020)Acemoglu, Autor, Hazell, and
  Restrepo]{acemoglu2020ai}
Daron Acemoglu, David Autor, Jonathon Hazell, and Pascual Restrepo.
\newblock Ai and jobs: Evidence from online vacancies.
\newblock Working Paper 28257, National Bureau of Economic Research, December
  2020.
\newblock URL \url{http://www.nber.org/papers/w28257}.

\bibitem[Agrawal et~al.(2021)Agrawal, Gans, and Goldfarb]{agrawal2021ai}
Ajay~K Agrawal, Joshua~S Gans, and Avi Goldfarb.
\newblock Ai adoption and system-wide change.
\newblock Working Paper 28811, National Bureau of Economic Research, May 2021.
\newblock URL \url{http://www.nber.org/papers/w28811}.

\bibitem[Autor et~al.(2022)Autor, Mindell, Reynolds, and Solow]{autor2022work}
D.H. Autor, D.A. Mindell, E.~Reynolds, and R.M. Solow.
\newblock \emph{The Work of the Future: Building Better Jobs in an Age of
  Intelligent Machines}.
\newblock MIT Press, 2022.
\newblock ISBN 9780262046367.
\newblock URL \url{https://books.google.com/books?id=3tKMEAAAQBAJ}.

\bibitem[Metaxa et~al.(2021)Metaxa, Park, Robertson, Karahalios, Wilson,
  Hancock, and Sandvig]{metaxa2021audit}
Danaë Metaxa, Joon~Sung Park, Ronald~E. Robertson, Karrie Karahalios, Christo
  Wilson, Jeff Hancock, and Christian Sandvig.
\newblock Auditing algorithms: Understanding algorithmic systems from the
  outside in.
\newblock \emph{Foundations and Trends® in Human–Computer Interaction},
  14\penalty0 (4):\penalty0 272--344, 2021.
\newblock ISSN 1551-3955.
\newblock \doi{10.1561/1100000083}.
\newblock URL \url{http://dx.doi.org/10.1561/1100000083}.

\bibitem[Raji and Buolamwini(2019)]{raji2019actionable}
Inioluwa~Deborah Raji and Joy Buolamwini.
\newblock Actionable auditing: Investigating the impact of publicly naming
  biased performance results of commercial ai products.
\newblock In \emph{Proceedings of the 2019 AAAI/ACM Conference on AI, Ethics,
  and Society}, AIES '19, page 429–435, New York, NY, USA, 2019. Association
  for Computing Machinery.
\newblock ISBN 9781450363242.
\newblock \doi{10.1145/3306618.3314244}.
\newblock URL \url{https://doi.org/10.1145/3306618.3314244}.

\bibitem[Kleinberg and Raghavan(2021)]{kleinberg2021monoculture}
Jon Kleinberg and Manish Raghavan.
\newblock Algorithmic monoculture and social welfare.
\newblock \emph{Proceedings of the National Academy of Sciences}, 118\penalty0
  (22):\penalty0 e2018340118, 2021.
\newblock \doi{10.1073/pnas.2018340118}.
\newblock URL \url{https://www.pnas.org/doi/abs/10.1073/pnas.2018340118}.

\bibitem[Bommasani et~al.(2022)Bommasani, Creel, Kumar, Jurafsky, and
  Liang]{bommasani2022homogenization}
Rishi Bommasani, Kathleen~A. Creel, Ananya Kumar, Dan Jurafsky, and Percy
  Liang.
\newblock Picking on the same person: Does algorithmic monoculture lead to
  outcome homogenization?
\newblock In \emph{Advances in Neural Information Processing Systems}, 2022.

\bibitem[Deng et~al.(2009)Deng, Dong, Socher, Li, Li, and
  Fei-Fei]{deng2009imagenet}
Jia Deng, Wei Dong, Richard Socher, Li-Jia Li, Kai Li, and Li~Fei-Fei.
\newblock {I}mage{N}et: A large-scale hierarchical image database.
\newblock In \emph{Computer Vision and Pattern Recognition (CVPR)}, pages
  248--255, 2009.

\bibitem[Ribeiro et~al.(2020)Ribeiro, Wu, Guestrin, and
  Singh]{ribeiro2020beyond}
Marco~Tulio Ribeiro, Tongshuang Wu, Carlos Guestrin, and Sameer Singh.
\newblock Beyond accuracy: Behavioral testing of {NLP} models with
  {C}heck{L}ist.
\newblock In \emph{Association for Computational Linguistics (ACL)}, pages
  4902--4912, 2020.

\bibitem[Perez et~al.(2022)Perez, Huang, Song, Cai, Ring, Aslanides, Glaese,
  McAleese, and Irving]{perez2022red}
Ethan Perez, Saffron Huang, Francis Song, Trevor Cai, Roman Ring, John
  Aslanides, Amelia Glaese, Nathan McAleese, and Geoffrey Irving.
\newblock Red teaming language models with language models.
\newblock \emph{ArXiv}, abs/2202.03286, 2022.

\bibitem[Bommasani and Cardie(2020)]{bommasani2020intrinsic}
Rishi Bommasani and Claire Cardie.
\newblock Intrinsic evaluation of summarization datasets.
\newblock In \emph{Proceedings of the 2020 Conference on Empirical Methods in
  Natural Language Processing (EMNLP)}, pages 8075--8096, Online, November
  2020. Association for Computational Linguistics.
\newblock \doi{10.18653/v1/2020.emnlp-main.649}.
\newblock URL \url{https://aclanthology.org/2020.emnlp-main.649}.

\bibitem[Swayamdipta et~al.(2020)Swayamdipta, Schwartz, Lourie, Wang,
  Hajishirzi, Smith, and Choi]{swayamdipta2020dataset}
Swabha Swayamdipta, Roy Schwartz, Nicholas Lourie, Yizhong Wang, Hannaneh
  Hajishirzi, Noah~A. Smith, and Yejin Choi.
\newblock Dataset cartography: Mapping and diagnosing datasets with training
  dynamics.
\newblock In \emph{Empirical Methods in Natural Language Processing (EMNLP)},
  2020.

\bibitem[Ethayarajh et~al.(2022)Ethayarajh, Choi, and
  Swayamdipta]{ethayarajh2022dataset}
Kawin Ethayarajh, Yejin Choi, and Swabha Swayamdipta.
\newblock Understanding dataset difficulty with $\mathcal{V}$-usable
  information.
\newblock In Kamalika Chaudhuri, Stefanie Jegelka, Le~Song, Csaba Szepesvari,
  Gang Niu, and Sivan Sabato, editors, \emph{Proceedings of the 39th
  International Conference on Machine Learning}, volume 162 of
  \emph{Proceedings of Machine Learning Research}, pages 5988--6008. PMLR,
  17--23 Jul 2022.
\newblock URL \url{https://proceedings.mlr.press/v162/ethayarajh22a.html}.

\bibitem[Mitchell et~al.(2022)Mitchell, Luccioni, Lambert, Gerchick,
  McMillan-Major, Ozoani, Rajani, Thrush, Jernite, and
  Kiela]{mitchell2022measuring}
Margaret Mitchell, Alexandra~Sasha Luccioni, Nathan Lambert, Marissa Gerchick,
  Angelina McMillan-Major, Ezinwanne Ozoani, Nazneen Rajani, Tristan Thrush,
  Yacine Jernite, and Douwe Kiela.
\newblock Measuring data.
\newblock \emph{ArXiv}, abs/2212.05129, 2022.

\bibitem[Lee et~al.(2022{\natexlab{a}})Lee, Liang, and Yang]{lee2022coauthor}
Mina Lee, Percy Liang, and Qian Yang.
\newblock Coauthor: Designing a human-{AI} collaborative writing dataset for
  exploring language model capabilities.
\newblock In \emph{Conference on Human Factors in Computing Systems (CHI)},
  2022{\natexlab{a}}.

\bibitem[Lee et~al.(2022{\natexlab{b}})Lee, Srivastava, Hardy, Thickstun,
  Durmus, Paranjape, Gerard-Ursin, Li, Ladhak, Rong, Wang, Kwon, Park, Cao,
  Lee, Bommasani, Bernstein, and Liang]{lee2022interaction}
Mina Lee, Megha Srivastava, Amelia Hardy, John Thickstun, Esin Durmus, Ashwin
  Paranjape, Ines Gerard-Ursin, Xiang~Lisa Li, Faisal Ladhak, Frieda Rong,
  Rose~E. Wang, Minae Kwon, Joon~Sung Park, Hancheng Cao, Tony Lee, Rishi
  Bommasani, Michael Bernstein, and Percy Liang.
\newblock Evaluating human-language model interaction.
\newblock 2022{\natexlab{b}}.
\newblock URL \url{https://arxiv.org/abs/2212.09746}.

\bibitem[Crisan et~al.(2022)Crisan, Drouhard, Vig, and
  Rajani]{crisan2022interactive}
Anamaria Crisan, Margaret Drouhard, Jesse Vig, and Nazneen Rajani.
\newblock Interactive model cards: A human-centered approach to model
  documentation.
\newblock In \emph{2022 ACM Conference on Fairness, Accountability, and
  Transparency}, FAccT '22, page 427–439, New York, NY, USA, 2022.
  Association for Computing Machinery.
\newblock ISBN 9781450393522.
\newblock \doi{10.1145/3531146.3533108}.
\newblock URL \url{https://doi.org/10.1145/3531146.3533108}.

\bibitem[Dotan and Milli(2020)]{dotan2020value}
Ravit Dotan and Smitha Milli.
\newblock Value-laden disciplinary shifts in machine learning.
\newblock \emph{Proceedings of the 2020 Conference on Fairness, Accountability,
  and Transparency}, 2020.

\bibitem[Ethayarajh and Jurafsky(2020)]{ethayarajh2020utility}
Kawin Ethayarajh and Dan Jurafsky.
\newblock Utility is in the eye of the user: A critique of {NLP} leaderboards.
\newblock In \emph{Proceedings of the 2020 Conference on Empirical Methods in
  Natural Language Processing (EMNLP)}, pages 4846--4853, Online, November
  2020. Association for Computational Linguistics.
\newblock \doi{10.18653/v1/2020.emnlp-main.393}.
\newblock URL \url{https://aclanthology.org/2020.emnlp-main.393}.

\bibitem[Scheuerman et~al.(2021)Scheuerman, Hanna, and
  Denton]{scheuerman2021politics}
Morgan~Klaus Scheuerman, Alex Hanna, and Emily Denton.
\newblock Do datasets have politics? disciplinary values in computer vision
  dataset development.
\newblock \emph{Proc. ACM Hum.-Comput. Interact.}, 5\penalty0 (CSCW2), oct
  2021.
\newblock \doi{10.1145/3476058}.
\newblock URL \url{https://doi.org/10.1145/3476058}.

\bibitem[Raji et~al.(2021)Raji, Denton, Bender, Hanna, and
  Paullada]{raji2021ai}
Inioluwa~Deborah Raji, Emily Denton, Emily~M. Bender, Alex Hanna, and
  Amandalynne Paullada.
\newblock {AI} and the everything in the whole wide world benchmark.
\newblock In \emph{Thirty-fifth Conference on Neural Information Processing
  Systems Datasets and Benchmarks Track (Round 2)}, 2021.
\newblock URL \url{https://openreview.net/forum?id=j6NxpQbREA1}.

\bibitem[Koch et~al.(2021)Koch, Denton, Hanna, and Foster]{koch2021reduced}
Bernard Koch, Emily Denton, Alex Hanna, and Jacob~Gates Foster.
\newblock Reduced, reused and recycled: The life of a dataset in machine
  learning research.
\newblock In \emph{Thirty-fifth Conference on Neural Information Processing
  Systems Datasets and Benchmarks Track (Round 2)}, 2021.
\newblock URL \url{https://openreview.net/forum?id=zNQBIBKJRkd}.

\bibitem[Denton et~al.(2021)Denton, Hanna, Amironesei, Smart, and
  Nicole]{denton2021genealogy}
Emily Denton, Alex Hanna, Razvan Amironesei, Andrew Smart, and Hilary Nicole.
\newblock On the genealogy of machine learning datasets: A critical history of
  imagenet.
\newblock \emph{Big Data \& Society}, 8\penalty0 (2):\penalty0
  20539517211035955, 2021.
\newblock \doi{10.1177/20539517211035955}.
\newblock URL \url{https://doi.org/10.1177/20539517211035955}.

\bibitem[Bommasani(2022)]{bommasani2022evaluation}
Rishi Bommasani.
\newblock Evaluation for change.
\newblock \emph{ArXiv}, 2022.
\newblock URL \url{https://arxiv.org/abs/2212.11670}.

\bibitem[Lazar(2023)]{lazar2023algorithmiccity}
Seth Lazar.
\newblock Governing the algorithmic city.
\newblock \emph{Tanner Lectures}, 2023.
\newblock URL \url{https://write.as/sethlazar/}.

\bibitem[Wang et~al.(2022{\natexlab{b}})Wang, Kapoor, Barocas, and
  Narayanan]{wang2022against}
Angelina Wang, Sayash Kapoor, Solon Barocas, and Arvind Narayanan.
\newblock Against predictive optimization: On the legitimacy of decision-making
  algorithms that optimize predictive accuracy.
\newblock \emph{Available at SSRN}, 2022{\natexlab{b}}.

\bibitem[Stone et~al.(2022)Stone, Brooks, Brynjolfsson, Calo, Etzioni, Hager,
  Hirschberg, Kalyanakrishnan, Kamar, Kraus, et~al.]{stone2022artificial}
Peter Stone, Rodney Brooks, Erik Brynjolfsson, Ryan Calo, Oren Etzioni, Greg
  Hager, Julia Hirschberg, Shivaram Kalyanakrishnan, Ece Kamar, Sarit Kraus,
  et~al.
\newblock Artificial intelligence and life in 2030: the one hundred year study
  on artificial intelligence.
\newblock \emph{arXiv preprint arXiv:2211.06318}, 2022.

\bibitem[Zhang et~al.(2021)Zhang, Mishra, Brynjolfsson, Etchemendy, Ganguli,
  Grosz, Lyons, Manyika, Niebles, Sellitto, et~al.]{zhang2021ai}
Daniel Zhang, Saurabh Mishra, Erik Brynjolfsson, John Etchemendy, Deep Ganguli,
  Barbara Grosz, Terah Lyons, James Manyika, Juan~Carlos Niebles, Michael
  Sellitto, et~al.
\newblock The ai index 2021 annual report.
\newblock \emph{arXiv preprint arXiv:2103.06312}, 2021.

\bibitem[Wolf et~al.(2020)Wolf, Debut, Sanh, Chaumond, Delangue, Moi, Cistac,
  Rault, Louf, Funtowicz, Davison, Shleifer, von Platen, Ma, Jernite, Plu, Xu,
  Le~Scao, Gugger, Drame, Lhoest, and Rush]{wolf2020transformers}
Thomas Wolf, Lysandre Debut, Victor Sanh, Julien Chaumond, Clement Delangue,
  Anthony Moi, Pierric Cistac, Tim Rault, Remi Louf, Morgan Funtowicz, Joe
  Davison, Sam Shleifer, Patrick von Platen, Clara Ma, Yacine Jernite, Julien
  Plu, Canwen Xu, Teven Le~Scao, Sylvain Gugger, Mariama Drame, Quentin Lhoest,
  and Alexander Rush.
\newblock Transformers: State-of-the-art natural language processing.
\newblock In \emph{Proceedings of the 2020 Conference on Empirical Methods in
  Natural Language Processing: System Demonstrations}, pages 38--45, Online,
  October 2020. Association for Computational Linguistics.
\newblock \doi{10.18653/v1/2020.emnlp-demos.6}.
\newblock URL \url{https://aclanthology.org/2020.emnlp-demos.6}.

\bibitem[Lhoest et~al.(2021)Lhoest, del Moral, Jernite, Thakur, von Platen,
  Patil, Chaumond, Drame, Plu, Tunstall, Davison, vSavsko, Chhablani, Malik,
  Brandeis, Scao, Sanh, Xu, Patry, McMillan-Major, Schmid, Gugger, Delangue,
  Matussiere, Debut, Bekman, Cistac, Goehringer, Mustar, Lagunas, Rush, and
  Wolf]{lhoest2021datasets}
Quentin Lhoest, Albert~Villanova del Moral, Yacine Jernite, Abhishek Thakur,
  Patrick von Platen, Suraj Patil, Julien Chaumond, Mariama Drame, Julien Plu,
  Lewis Tunstall, Joe Davison, Mario vSavsko, Gunjan Chhablani, Bhavitvya
  Malik, Simon Brandeis, Teven~Le Scao, Victor Sanh, Canwen Xu, Nicolas Patry,
  Angelina McMillan-Major, Philipp Schmid, Sylvain Gugger, Clement Delangue,
  Th'eo Matussiere, Lysandre Debut, Stas Bekman, Pierric Cistac, Thibault
  Goehringer, Victor Mustar, François Lagunas, Alexander~M. Rush, and Thomas
  Wolf.
\newblock Datasets: A community library for natural language processing.
\newblock \emph{ArXiv}, abs/2109.02846, 2021.

\bibitem[{White House Executive Order}(2021)]{whitehouse2021cybersecurity}
{White House Executive Order}.
\newblock Executive order on improving the nation’s cybersecurity.
\newblock 2021.
\newblock URL
  \url{https://www.whitehouse.gov/briefing-room/presidential-actions/2021/05/12/executive-order-on-improving-the-nations-cybersecurity/}.

\bibitem[Ramaswami(2021)]{ramaswami2021securing}
Ashwin Ramaswami.
\newblock Securing open source software at the source.
\newblock \emph{Plaintext Group by Schmidt Futures}, 2021.
\newblock URL
  \url{https://www.plaintextgroup.com/reports/securing-open-source-software-at-the-source}.

\bibitem[Scott et~al.(2023)Scott, Brackett, Herr, and
  Hamin]{scott2023atlanticcouncil}
Stewart Scott, Sara~Ann Brackett, Trey Herr, and Maia Hamin.
\newblock Avoiding the success trap: Toward policy for open-source software as
  infrastructure.
\newblock \emph{Atlantic Council}, 2023.
\newblock URL
  \url{https://www.atlanticcouncil.org/in-depth-research-reports/report/open-source-software-as-infrastructure/}.

\bibitem[Rowlinson(1997)]{rowlinson1997organisations}
Michael Rowlinson.
\newblock \emph{Organisations and institutions: perspectives in economics and
  sociology}.
\newblock Springer, 1997.

\bibitem[Weingast and Marshall(1988)]{weingast1988industrial}
Barry~R. Weingast and William~J. Marshall.
\newblock The industrial organization of congress; or, why legislatures, like
  firms, are not organized as markets.
\newblock \emph{Journal of Political Economy}, 96\penalty0 (1):\penalty0
  132--163, 1988.
\newblock \doi{10.1086/261528}.
\newblock URL \url{https://doi.org/10.1086/261528}.

\bibitem[Einav and Levin(2010)]{einav2010empirical}
Liran Einav and Jonathan Levin.
\newblock Empirical industrial organization: A progress report.
\newblock \emph{Journal of Economic Perspectives}, 24\penalty0 (2):\penalty0
  145--62, June 2010.
\newblock \doi{10.1257/jep.24.2.145}.
\newblock URL \url{https://www.aeaweb.org/articles?id=10.1257/jep.24.2.145}.

\bibitem[Frickel and Moore(2006)]{frickel2006new}
Scott Frickel and Kelly Moore.
\newblock \emph{The new political sociology of science: Institutions, networks,
  and power}.
\newblock Univ of Wisconsin Press, 2006.

\bibitem[Dequech(2006)]{dequech2006institutions}
David Dequech.
\newblock Institutions and norms in institutional economics and sociology.
\newblock \emph{Journal of Economic Issues}, 40\penalty0 (2):\penalty0
  473--481, 2006.

\bibitem[Fleury(2014)]{fleury2014sociology}
Laurent Fleury.
\newblock \emph{Sociology of culture and cultural practices: The transformative
  power of institutions}.
\newblock Lexington Books, 2014.

\end{thebibliography}

\section*{Checklist}

\begin{enumerate}

\item For all authors...
\begin{enumerate}
  \item Do the main claims made in the abstract and introduction accurately reflect the paper's contributions and scope?
    \answerYes{}
  \item Did you describe the limitations of your work?
    \answerYes{}
  \item Did you discuss any potential negative societal impacts of your work?
    \answerNo{We do not foresee serious negative societal impacts, beyond misinterpretation stemming from the limitations we do discuss.}
  \item Have you read the ethics review guidelines and ensured that your paper conforms to them?
    \answerYes{}
\end{enumerate}

\item If you are including theoretical results...
\begin{enumerate}
  \item Did you state the full set of assumptions of all theoretical results?
    \answerNA{}
        \item Did you include complete proofs of all theoretical results?
    \answerNA{}
\end{enumerate}

\item If you ran experiments...
\begin{enumerate}
  \item Did you include the code, data, and instructions needed to reproduce the main experimental results (either in the supplemental material or as a URL)?
    \answerNA{}
  \item Did you specify all the training details (e.g., data splits, hyperparameters, how they were chosen)?
    \answerNA{}
        \item Did you report error bars (e.g., with respect to the random seed after running experiments multiple times)?
    \answerNA{}
        \item Did you include the total amount of compute and the type of resources used (e.g., type of GPUs, internal cluster, or cloud provider)?
    \answerNA{}
\end{enumerate}

\item If you are using existing assets (e.g., code, data, models) or curating/releasing new assets...
\begin{enumerate}
  \item If your work uses existing assets, did you cite the creators?
    \answerYes{} We directly point to the link associated with the asset and foreground the asset developer's organization.
  \item Did you mention the license of the assets?
    \answerYes{} Licenses are a field in our annotation schema.
  \item Did you include any new assets either in the supplemental material or as a URL?
    \answerNA{}
  \item Did you discuss whether and how consent was obtained from people whose data you're using/curating?
    \answerYes{The data is public information that is centralized from standard sources (\eg papers, blog posts, news and journalism).}
  \item Did you discuss whether the data you are using/curating contains personally identifiable information or offensive content?
    \answerNA{}
\end{enumerate}

\item If you used crowdsourcing or conducted research with human subjects...
\begin{enumerate}
  \item Did you include the full text of instructions given to participants and screenshots, if applicable?
    \answerNA{}
  \item Did you describe any potential participant risks, with links to Institutional Review Board (IRB) approvals, if applicable?
    \answerNA{}
  \item Did you include the estimated hourly wage paid to participants and the total amount spent on participant compensation?
    \answerNA{}
\end{enumerate}

\end{enumerate}



\end{document}